\documentclass[lettersize,journal]{IEEEtran}
\usepackage{cite}
\usepackage{graphicx}
\usepackage{amsmath,amssymb,amsfonts}
\usepackage{array}          
\usepackage{booktabs}       
\usepackage{multirow}       

\usepackage[caption=false,font=footnotesize,labelfont=sf,textfont=sf]{subfig}
\usepackage{caption} 
\usepackage{tabularx}
\usepackage{pifont}
\usepackage{adjustbox}
\usepackage{enumitem}
\usepackage{soul}

\usepackage{amsthm}
\theoremstyle{definition}
\newtheorem{definition}{Definition}[section]

\usepackage{tikz}
\usepackage{url} 
\usetikzlibrary{positioning, arrows.meta, shapes, fit, calc}
\usepackage{hyperref}

\title{Mid-Training of Large Language Models: A Survey}

\author{Kaixiang Mo, Yuxin Shi, Weiwei Weng, Zhiqiang Zhou, Shuman Liu, Haibo Zhang$^\dagger$, Anxiang Zeng$^\dagger$
\thanks{All authors are with Shopee. E-mails: \{kaixiang.mo, yuxin.shi, weiwei.weng, zhiqiang.zhou, shuman.liu, peter.wu\}@shopee.com, zeng0118@ntu.edu.sg}
\thanks{$^\dagger$Corresponding author: peter.wu@shopee.com, zeng0118@ntu.edu.sg}
\thanks{This work has been submitted to the IEEE for possible publication.}}

\begin{document}

\maketitle
\bstctlcite{BSTcontrol}

\begin{abstract}
Large language models (LLMs) are typically developed through large-scale pre-training followed by task-specific fine-tuning. Recent advances highlight the importance of an intermediate mid-training stage, where models undergo multiple annealing-style phases that refine data quality, adapts optimization schedules, and extend context length. This stage mitigates diminishing returns from noisy tokens, stabilizes convergence, and expands model capability in late training. Its effectiveness can be explained through gradient noise scale, the information bottleneck, and curriculum learning, which together promote generalization and abstraction. Despite widespread use in state-of-the-art systems, there has been no prior survey of mid-training as a unified paradigm. 
We introduce the first taxonomy of LLM mid-training spanning data distribution, learning-rate scheduling, and long-context extension. We distill practical insights, compile evaluation benchmarks, and report gains to enable structured comparisons across models. We also identify open challenges and propose avenues for future research and practice.

\end{abstract}
\begin{figure*}
    \centering
    \captionsetup{justification=raggedright,singlelinecheck=false}
    \includegraphics[width=1\linewidth]{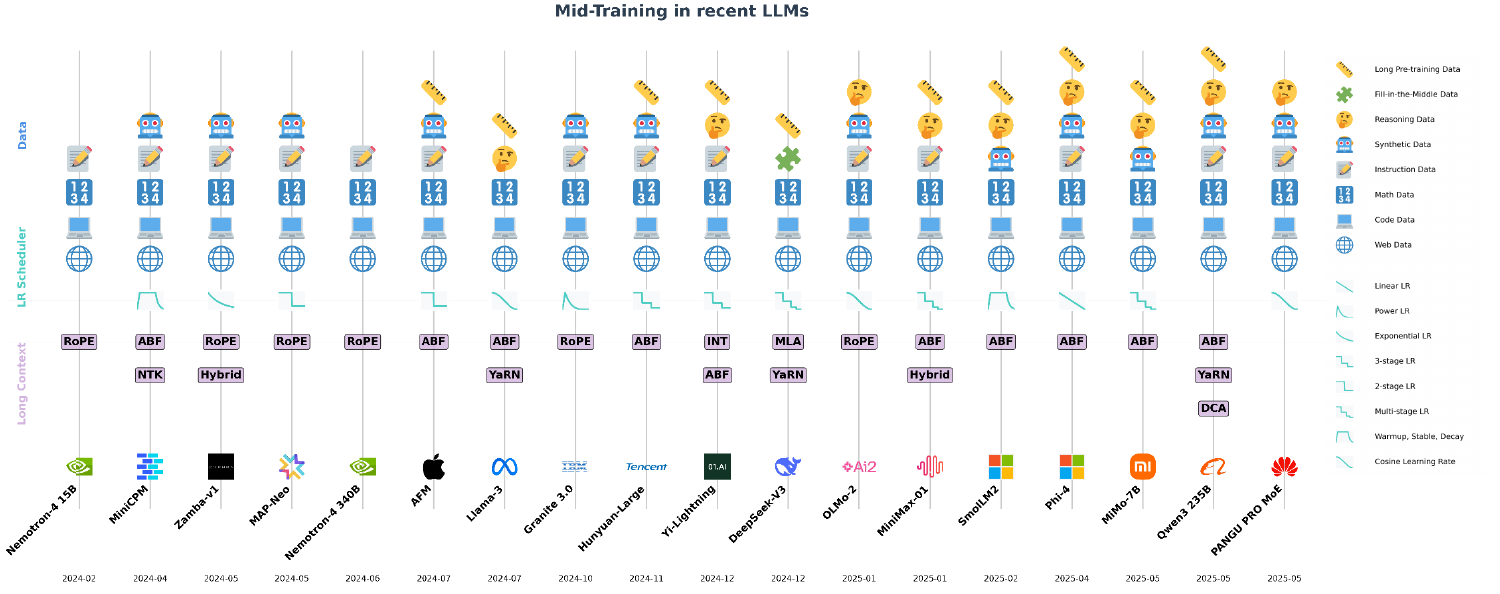}
    \caption{Mid Training of recent models. Only selected models with detailed data mix are shown. In the long context part, Hybrid means a combination of different attention mechanisms, e.g., self-attention, SSM, lightning-attention, etc. INT is the abbreviation for interleaving global and local attentions. }
    \label{fig:data_lr_longcontext}
\end{figure*}

\section{Introduction}
\label{sec:intro}

In recent large language model (LLM) training, mid-training has emerged as a crucial continuation stage between general pre-training and task-specific supervised fine-tuning (SFT). Rather than relying solely on web-scale corpora, which provide broad but noisy supervision, state-of-the-art models increasingly adopt an annealing-style phase to refine optimization dynamics and sharpen data quality~\cite{dubey2024llama, hu2024minicpm, glorioso2024zamba}. This stage is motivated by both practical and theoretical considerations: lowering the learning rate mitigates gradient variance and stabilizes convergence near favorable minima, while shifting toward more curated or synthetic corpora~\cite{abdin2024phi3} enhances the marginal utility of each additional token. Without such adjustments, models often stagnate in capability development, showing limited gains in reasoning, coding, and long-context understanding despite substantial additional compute~\cite{bi2024deepseek, allal2025smollm2}. 

The effectiveness of mid-training can be theoretically motivated, as it shifts models from memorization toward abstraction by emphasizing structured reasoning, factuality, and instruction following. We summarize this phenomenon from three perspectives: (i) gradient noise scale, where high-quality data improves optimization by enhancing signal variance and mitigating overfitting plateaus~\cite{mccandlish2018empirical, wang2021eliminating, meng2023per}; (ii) the information bottleneck, which compresses noisy features while preserving predictive structures~\cite{izmailov2022feature}; and (iii) curriculum learning, where data distributions are gradually refined to reinforce complex reasoning~\cite{wang2025dump, jia2022sample}. Together, these perspectives provide a principled rationale for mid-training as a strategy to improve generalization and efficiency in large-scale training.

Despite its growing adoption and effectiveness, there has been no comprehensive survey that conceptualizes mid-training as a coherent paradigm. Existing studies have explored isolated aspects (e.g., curriculum-style data annealing~\cite{du2024survey,wu2025data}, adaptive learning-rate schedules~\cite{jin2023rethinking}, or context-length extension~\cite{huang2023advancing, liu2025comprehensive, pawar2024and}) but lack a comprehensive framework. To bridge this gap, we present an integrated review of mid-training strategies across three interconnected domains: data distribution, optimization scheduling, and context extension. By framing these elements as mutually reinforcing, we highlight mid-training as a distinct and coherent stage in the LLM development pipeline, rather than an ad-hoc collection of heuristics.

A central aspect of mid-training is the refinement of the data mixture after large-scale pre-training. At this point, models have acquired broad linguistic and semantic features from trillions of web-scale tokens, making the marginal value of additional noisy data limited. To improve efficiency, the distribution is shifted toward curated or synthetic corpora emphasizing reasoning, coding, STEM, multilinguality, and instruction following. Empirical evidence shows that such targeted data can foster compositional skills (e.g., chain-of-thought reasoning, math problem solving, and multi-step planning) that are underrepresented in general crawls~\cite{abdin2024phi, allal2025smollm2}. Common practice includes down-sampling low-quality tokens and up-sampling knowledge-dense corpora, often through iterative learning rate annealing passes~\cite{NEURIPS2024_19e4ea30, penedo2024fineweb, Su2024NemotronCCTC}. 

Another key component of mid-training is adapting the learning rate schedule. After aggressive pre-training, models approach promising optima where large step sizes risk instability or divergence. Reducing the learning rate with smoother decay stabilizes convergence and suppresses gradient noise~\cite{you1908does, gilmer2022loss, kalra2024warmup}. Existing strategies include linear or cosine decay, multi-stage schedulers, and adaptive schemes tailored to long runs~\cite{goyal2017accurate, loshchilov2016sgdr, bi2024deepseek, hu2024minicpm, shen2024power, ibrahim2024simple}. These methods improve sample efficiency by enabling finer assimilation of high-quality tokens. Yet scheduler design remains largely empirical: the optimal duration and shape of decay vary across model sizes, architectures, and optimizers~\cite{kaplan2020scaling, hoffmann2022training, defazio2023optimal, bergsma2025straight, li2025predictable}. Overly steep decay may halt learning prematurely, while overly cautious schedules waste computation with limited gains.

Finally, mid-training is also a natural stage for extending context length beyond the 4K–8K limits of early pre-training. This extension supports downstream tasks that demand coherence across long documents, multi-file reasoning, or dialogue histories spanning tens of thousands of tokens. Current methods combine positional encoding remapping, such as Position Interpolation~\cite{chen2023extending}, NTK-aware interpolation~\cite{bloc97_ntk_rope_2023}, YaRN~\cite{peng2023yarn} or ABF~\cite{xiong2023effective} with curricula that gradually introduce long-form inputs. Studies show that extending context during mid-training enables models to capture long-range dependencies more effectively without restarting pre-training. 

Together, these elements form the backbone of mid-training: without refined data, the model lacks specialization; without proper scheduling, it risks instability or underutilization of valuable tokens; and without context extension, it remains constrained in scope for downstream applications. Their interdependence suggests that effective mid-training cannot be understood by analyzing each factor in isolation, but rather through their combined impact on model performance and efficiency. This holistic view motivates the need for systematic discussion of mid-training as a distinct stage in the LLM development pipeline.

\textbf{Our contributions}. This paper contributes to academia and industry in the following ways:
\begin{itemize}[leftmargin=10pt]
    \item We propose a taxonomy of LLM mid-training based on three domains (data distribution, learning rate scheduling and long-context extension), and summarize various approaches in each domain. To the best of our knowledge, it is the first such taxonomy on LLM mid-training.
    \item We summarize valuable insights in each domain, providing readers with a convenient reference for future use.
    \item We discuss the common evaluation benchmarks and collect the reported gains from different models, thereby, providing a structured comparison of how mid-training contribute to LLM training.
    \item We outline promising future research direction in LLM mid-training and propose potential ways forward.
\end{itemize}

\section{Possible Theory Behind Mid-Training}
The core rationale for mid-training is to shift the model's learning dynamics from memorization to abstraction. As the model approaches the capacity limits of broad generalization from large-scale general or medium-quality data, high-quality samples enable further gains by sharpening attention to structured reasoning, factuality, and instruction following. Below, we outline several theoretical foundations for this approach.

\begin{table*}[h]
\centering
\scriptsize
\renewcommand{\arraystretch}{1.2}
\newcolumntype{L}[1]{>{\raggedright\arraybackslash}p{#1}}
\caption{Consolidated overview of state-of-the-art models that disclose their mid-training datasets.}
\label{tab:open_sourced_ds}

\begin{adjustbox}{width=\textwidth}
\begin{tabularx}{\textwidth}{@{} l L{3cm} L{2.5cm} L{3cm} L{3cm} L{3cm} @{}}
\toprule
\textbf{Model} & \textbf{HQ Web Data} & \textbf{Code} & \textbf{Math} & \textbf{Instruction\& QA Style} & \textbf{Synthetic Data} \\
\midrule
\textbf{MiniCPM}\mbox{\cite{hu2024minicpm}} & CommonCrawl\_Chn; Dolma~\mbox{\cite{soldaini2024dolma}}; C4~\mbox{\cite{C4dataset}}; Pile~\mbox{\cite{gao2020pile}}; Wikipedia~\mbox{\cite{wikidump}}; Arxiv~\mbox{\cite{clement2019usearxivdataset}}; BaiduBaike & Stack~\mbox{\cite{Kocetkov2022TheStack}} & OpenWebMath~\mbox{\cite{paster2023openwebmath}} & SlimOrca~\mbox{\cite{SlimOrca}}; OssInstruct~\mbox{\cite{wei2024magicoder}}; UltraChat~\mbox{\cite{ding2023enhancing}}; EvolInstruct~\mbox{\cite{surge2024openbezoar}}; Stack Exchange QA & \textemdash \\
\textbf{Zamba-v1}\mbox{\cite{glorioso2024zamba}} & Pile~\mbox{\cite{gao2020pile}}; C4~\mbox{\cite{C4dataset}}; PeS2o~\mbox{\cite{peS2o}}; Arxiv; RefinedWeb~\mbox{\cite{penedo2023refinedweb}} & EvolInstructCode\mbox{\cite{luo2023wizardcoder}} & StackMathQA~\mbox{\cite{stackmathqa2024}} & OpenOrca~\mbox{\cite{OpenOrca}} & Cosmopedia~\mbox{\cite{benallal2024cosmopedia}} \\
\textbf{MAP-Neo}\mbox{\cite{Zhang2024MAPNeoHC}} & Matrix~\mbox{\cite{Zhang2024MAPNeoHC}}; US-PD-Books~\mbox{\cite{US-PD-Books}} & Matrix~\mbox{\cite{Zhang2024MAPNeoHC}}; Stack V2~\mbox{\cite{lozhkov2024starcoder}} & Matrix~\mbox{\cite{Zhang2024MAPNeoHC}}; AutoMathText~\mbox{\cite{zhang2025autonomous}}; OpenWebMath~\mbox{\cite{paster2023openwebmath}} & Matrix~\mbox{\cite{Zhang2024MAPNeoHC}}; BioInstrutQA~\mbox{\cite{BioInstructQA}}; SMolInstruct~\mbox{\cite{yu2024llasmol}}; Agent-FLAN~\mbox{\cite{chen2024agent}} & Matrix~\mbox{\cite{Zhang2024MAPNeoHC}}; Cosmopedia~\mbox{\cite{benallal2024cosmopedia}} \\
\textbf{OLMo-2}\mbox{\cite{olmo20242}} & 
Dolmino-Mix-1124~\mbox{\cite{olmo20242}} (DCLM-Baseline~\mbox{\cite{NEURIPS2024_19e4ea30}}; PeS2o-Dolma1.7~\mbox{\cite{soldaini2024dolma}}; Wiki-Dolma1.7~\mbox{\cite{soldaini2024dolma}}) & Dolmino-Mix-1124~\mbox{\cite{olmo20242}} (CodeSearchNet-OWMfilter~\mbox{\cite{husain2019codesearchnet}}) & Dolmino-Mix-1124~\mbox{\cite{olmo20242}} (Metamath-OWMfilter~\mbox{\cite{yu2023metamath}}; GSM8K-trainsplit~\mbox{\cite{cobbe2021gsm8k}}) & Dolmino-Mix-1124~\mbox{\cite{olmo20242}} (Flan-Dolma1.7~\mbox{\cite{soldaini2024dolma}}; StackExchange-RedPajamaV1~\mbox{\cite{together2023redpajama}}) & Dolmino-Mix-1124~\mbox{\cite{olmo20242}} (TuleMath; Dolmino SynthMath; TinyGSM-MIND; MathCoder2-synthetic)\\
\textbf{SmolLM2}\mbox{\cite{allal2025smollm2}} & DCLM-Baseline~\mbox{\cite{NEURIPS2024_19e4ea30}}; FineWeb-Edu~\mbox{\cite{penedo2024fineweb}} & Stack-Edu~\mbox{\cite{allal2025smollm2}} & InfiWebMath-3+~\mbox{\cite{han2024infimmwebmath40badvancingmultimodalpretraining}}; FineMath 4+~\mbox{\cite{allal2025smollm2}}; OpenWebMath~\mbox{\cite{paster2023openwebmath}}; AugGSM8K~\mbox{\cite{li2023mugglemath}} & \textemdash & Cosmopedia v2~\mbox{\cite{benallal2024smollmcorpus}} \\
\bottomrule
\end{tabularx}
\end{adjustbox}
\end{table*}

\noindent\textbf{Gradient Noise Scale:} 
Prior work shows that the gradient noise scale (GNS) reflects the amount of useful signal per update step~\cite{mccandlish2018empirical}. Higher-quality data tends to induce greater gradient variance, yielding a higher GNS, whereas redundant or noisy data reduces diversity. A larger GNS helps models escape sharp minima and avoid overfitting to low-quality plateaus~\cite{neelakantan2015adding, wang2021eliminating}. Empirically, this improves optimization in late training stages, where signal sparsity may stall convergence, and recent work shows that enhancing gradient signal-to-noise can further mitigate overfitting on noisy data~\cite{meng2023per}.

\noindent\textbf{Information Bottleneck:} 
In the information bottleneck (IB) framework, representation learning in neural networks can be interpreted as a process of compressing internal states while retaining task-relevant information~\cite{tishby2015deep}. During the learning rate annealing phase of training, the model progressively reduces reliance on noisy or redundant features. The model's representations may be viewed as increasingly emphasizing those features most predictive for downstream objectives~\cite{alemi2016deep}. In this view, high-quality supervision signals provide clearer, lower-entropy guidance that facilitates the identification of semantically meaningful structures and the attenuation of spurious correlations~\cite{izmailov2022feature}. Recent extensions of the IB principle to supervised settings further highlight how such compression can transform large-scale memorization into more abstract, generalizable representations~\cite{weingarten2025supervised}.

\noindent\textbf{Curriculum Learning:} 
Learning rate annealing naturally aligns with the principles of curriculum learning. Early pretraining exposes the model to diverse and noisy corpora for generalization, after which the data distribution can be gradually shifted toward more challenging and informative examples—exactly the role of curriculum-style scheduling~\cite{nair2024curriculum, chaudhry2024data}. This structured progression helps optimize learning efficiency and has been shown to reinforce complex skills such as multi-step reasoning and code generation~\cite{wang2025dump, jia2022sample}.

\section{Data Distributions}
\label{sec:data}
In this section, we will study the mid-training dataset distribution and quality in the state-of-the-art LLM approaches.

\subsection{Type of Commonly Used Mid-Training Data}
The mid-training stage in LLM pretraining represents a critical phase where data quality becomes increasingly prioritized to refine model capabilities, improve alignment, and optimize performance on downstream tasks. Across state-of-the-art LLMs, several distinct categories of data are commonly employed during mid-training. These can be broadly grouped into the following types:

\noindent\textbf{High-Quality Filtered Web Data:}
Models~\cite{hu2024minicpm, sun2024hunyuan, Dubey2024TheL3, allal2025smollm2, olmo20242, abdin2024phi} extensively utilize curated subsets of web-scale corpora. 
These subsets are meticulously filtered based on rigorous criteria including content quality, topic diversity, low toxicity, educational value, and minimal redundancy. In contrast to datasets used during initial pretraining stages, these refined datasets often undergo comprehensive filtering processes involving heuristic-based rules or learned scoring mechanisms to ensure selection of the most informative and high-quality samples.
Representative datasets include CommonCrawl, C4~\cite{C4dataset}, Wikipedia~\cite{wikidump}, Dolma~\cite{soldaini2024dolma}, RedPajama-Data-V2~\cite{together2023redpajama}, Culturax~\cite{nguyen2023culturax}, RefinedWeb~\cite{penedo2023refinedweb}, SlimPajama~\cite{shen2023slimpajama} and Matrix~\cite{Zhang2024MAPNeoHC}. Further specialized filtering has yielded higher-quality subsets such as FineWeb-Edu~\cite{penedo2024fineweb}, Ultra-FineWeb, DCLM-baseline~\cite{NEURIPS2024_19e4ea30}, peS2o~\cite{peS2o}.




\noindent\textbf{Code and Mathematical Content:}
Code and math data significantly enhance symbolic reasoning and program synthesis capabilities in mid-training stages~\cite{gunter2024apple, allal2025smollm2, Zhang2024MAPNeoHC, hu2024minicpm, chen2025minimax, yang2025qwen3, yang2025qwen3}.
Incorporating structured content from open-source repositories, curated coding benchmarks, and mathematical question-answer pairs or derivations enriches the model's ability to handle complex logic and technical reasoning.
Representative datasets include Stack v1~\cite{Kocetkov2022TheStack}, Stack v2~\cite{lozhkov2024starcoder}. More specifically filtered datasets include StackMathQA, OpenWebMath~\cite{paster2023openwebmath}, InfiMM-WebMath~\cite{han2024infimm}, FineMath~\cite{hu2024minicpm}, FineMath4+~\cite{hu2024minicpm}, FineMath3+~\cite{hu2024minicpm}, InfiWebMath~\cite{han2024infimmwebmath40badvancingmultimodalpretraining} , Infi-WebMath4+~\cite{hu2024minicpm}, Infi-WebMath3+~\cite{hu2024minicpm}, Stack-Edu~\cite{allal2025smollm2}, RefineCode~\cite{huang2024opencoder}, and EvolInstructCode~\cite{luo2023wizardcoder}.


\noindent\textbf{Instruction-Tuned and QA-Style Data:}
Instruction-following and QA-style data is increasingly prevalent during the mid-training~\cite{adler2024nemotron, hu2024minicpm, abdin2024phi, Zhang2024MAPNeoHC, sun2024hunyuan, olmo20242, chen2025minimax, tang2025pangu}. 
Such datasets typically comprise synthetic or curated question-answer pairs, instruction-response prompts, and alignment-tuning corpora. This type of data aims to enhance the model’s understanding of human intent, improve the consistency of the response, and strengthen the reasoning ability of the models~\cite{abdin2024phi}.
Representative datasets include Ultrachat~\cite{ding2023enhancing}, EvolInstruct~\cite{surge2024openbezoar}, OssInstruct~\cite{wei2024magicoder}, StackExchangeQA~\cite{olmo20242}, and FLAN~\cite{longpre2023flan}, OpenOrca~\cite{OpenOrca}, SMolInstruct~\cite{yu2024llasmol}.


\noindent\textbf{Synthetic Textbooks and Knowledge-Dense Data:}
Synthetic textbook-style and knowledge-dense data significantly enhance LLMs by providing high-quality educational content~\cite{abdin2024phi, allal2025smollm2, glorioso2024zamba, hu2024minicpm, sun2024hunyuan, gunter2024apple, olmo20242, yang2025qwen3}, particularly in low-resource domains such as math, coding, and multilingual contexts. These synthetic datasets are typically derived from knowledge graphs or generated using advanced LLMs by utilizing high-quality seeds sourced from multiple domains~\cite{abdin2024phi}. The synthetic data enrich the models with factual, explanatory, and pedagogically structured information. Such data strengthens the factual grounding and expands the world knowledge of models, improving their general reasoning and explanatory capabilities. 
Representative datasets include TuluMath~\cite{olmo20242}, Dolmino SynthMath~\cite{olmo20242}, TinyGSM-MIND~\cite{olmo20242}, MathCoder2 Synthetic~\cite{olmo20242}, Cosmopedia~\cite{benallal2024cosmopedia}, Cosmopedia v2~\cite{benallal2024smollmcorpus}, OpenHermes-2.5.

\noindent\textbf{Long-context Data:}
For models that support extended context lengths~\cite{yang2025qwen3, chen2025minimax, sun2024hunyuan, abdin2024phi}, the mid-training stage may incorporate long-context documents or long-form Q\&A designed to span thousands of tokens. 
These long-context data are either filtered from existing training corpus~\cite{sun2024hunyuan, abdin2024phi, allal2025smollm2, gao2024train} or obtained from data synthesis~\cite{abdin2024phi}. These samples improve the model’s memory, coherence, and reasoning across document-scale inputs.
Related datasets used to extract long-context data: DCLM~\cite{NEURIPS2024_19e4ea30}, FineWeb-Edu~\cite{penedo2024fineweb}, Dolma~\cite{soldaini2024dolma}

\noindent\textbf{Reasoning and CoT Data:}
Using reasoning data during mid-training has become a trend~\cite{tang2025pangu, yang2025qwen3}. Reasoning data, such as CoT annotations, provides explicit demonstrations of how to decompose complex problems into smaller steps, helping the model learn structured and interpretable reasoning patterns rather than relying only on surface-level cues. CoT reasoning data gives strong performance on math or logic tasks~\cite{sprague2024cot}. Such data is generally obtained from synthetic generation~\cite{abdin2024phi3, abdin2024phi} or through web data filtering. 

\noindent\textbf{Fill-in-Middle (FIM) Data:} Deepseek~\cite{DeepSeekAI2024DeepSeekV3TR, zhu2024deepseek} observes that the FIM strategy does not compromise the next-token prediction capability while enabling the model to accurately predict middle text based on contextual cues. Within the mid-training stage, FIM data is particularly beneficial: it enforces bi-directional contextual reasoning, strengthens long-range dependency modeling, and improves robustness under fragmented inputs. At the same time, FIM increases data efficiency by allowing multiple span-prediction opportunities from a single sample, making it well-suited for late-stage training where high-quality data is scarce. Similar observations have been made in code-pretrained models such as StarCoder~\cite{li2023starcoder} and CodeT5+~\cite{wang2023codet5+}, where infilling objectives significantly enhance both data utilization and downstream generalization.

Table.~\ref{tab:open_sourced_ds} offers a consolidated overview of state-of-the-art models that disclose their mid-training datasets, serving as a resource for the research community.

\begin{table*}[h]
\centering
\scriptsize
\caption{Structured summary of stages and data distributions employed during mid-training. For each stage, we indicate the number of phases and the total tokens consumed. In the case of the long-context stage, the number of phase reflects how many steps is used to progressively extend the sequence length.}
\label{fig:data_distribution_in_various_model}
\begin{adjustbox}{width=\textwidth}
\begin{tabularx}{\textwidth}{l| lll |*{8}{>{\centering\arraybackslash}X}}
\toprule
\textbf{Model} & \multicolumn{3}{c|}{\textbf{Stage Configuration (\# Phases, \# Tokens)}} & \multicolumn{8}{c}{\textbf{Data types (used in annealing \& long-context stages)}} \\
\cmidrule(lr){2-4} \cmidrule(lr){5-12}
& \textbf{General} & \textbf{Annealing} & \textbf{Long-context} & \textbf{HQ Web} & \textbf{Code} & \textbf{Math/ STEM} & \textbf{Instruct \& QA} & \textbf{Synthetic} & \textbf{Reasoning \& CoT} & \textbf{FIM} & \textbf{Long-context} \\

\midrule
Nemotron-4~\cite{adler2024nemotron} & 1 (8T) & 1 (1T) & - & \ding{51} & \ding{51} & \ding{51}  & \ding{51} &   &   &   &   \\
MiniCPM~\cite{hu2024minicpm} & 1 (1T) & 1 (20B) & - & \ding{51} & \ding{51} & \ding{51} & \ding{51} & \ding{51} &   &   &   \\
Phi-4~\cite{abdin2024phi} & 1 (10T) & 1 & 1 (250B) & \ding{51} & \ding{51} & \ding{51} & \ding{51} & \ding{51} & \ding{51} &  & \ding{51} \\
Deepseek-V3~\cite{DeepSeekAI2024DeepSeekV3TR} & 1 (14.8T) & - & 2 (120B) & \ding{51} & \ding{51} & \ding{51} &   &  &   & \ding{51} & \ding{51} \\
Zamba-v1~\cite{glorioso2024zamba} & 1 (950B) & 1 (1T) & - & \ding{51} & \ding{51} & \ding{51} & \ding{51} & \ding{51} &   &   &   \\
MAP-Neo~\cite{Zhang2024MAPNeoHC} & 1 (4.5T) & 1 (778B) & - & \ding{51} & \ding{51} & \ding{51} & \ding{51} & \ding{51} &   &   &   \\
AFM~\cite{gunter2024apple} & 1 (6.3T) & 1 (1T) & 1 (100B) & \ding{51} & \ding{51} & \ding{51} & \ding{51} & \ding{51} &   &   & \ding{51} \\
LLaMA3-405B~\cite{Dubey2024TheL3} & 1 (15T) & 1 (40B) & 6 (800B) & \ding{51} & \ding{51} & \ding{51} &   &   & \ding{51} &   & \ding{51} \\
Granite 3.0~\cite{granite2024granite} & 1 (8-10T) & 1 (2T) & - & \ding{51} & \ding{51} & \ding{51} & \ding{51} & \ding{51} &   &   &   \\
Hunyuan-Large~\cite{sun2024hunyuan} & 1 (7T) & 1 (350B) & 2 (10B) & \ding{51} & \ding{51} & \ding{51} & \ding{51} & \ding{51} &   &   & \ding{51} \\
Yi-Lightning~\cite{wake2024yi} & 1 & 2 & 3 (20B) & \ding{51} & \ding{51} & \ding{51} & \ding{51} &   & \ding{51} &   & \ding{51} \\
OLMo-2 13B~\cite{olmo20242} & 1 (5T) & 1 ($\sim$600B) & - & \ding{51} & \ding{51} & \ding{51} & \ding{51} & \ding{51} & \ding{51} &   &   \\
MiniMax-M1~\cite{chen2025minimax} & - & 1 (7.5T)\textsuperscript{*} & 3 (358B) & \ding{51} & \ding{51} & \ding{51} & \ding{51} &   & \ding{51} &   & \ding{51} \\
SmolLM2~\cite{allal2025smollm2} & 3 (10T) & 1 (1T) & 1 (75B) & \ding{51} & \ding{51} & \ding{51} &   & \ding{51} & \ding{51} &   & \ding{51} \\
Pangu Pro MoE~\cite{tang2025pangu} & 1 (9.6T) & 2 (3.4T, 32K)$^\dagger$ & - & \ding{51} & \ding{51} & \ding{51} & \ding{51} & \ding{51} & \ding{51} &   &   \\
Qwen3~\cite{yang2025qwen3} & 1 (30T) & 1 (5T) & 1 ($\sim$500B) & \ding{51} & \ding{51} & \ding{51} & \ding{51} & \ding{51} & \ding{51} &   & \ding{51}  \\
Mimo-7B~\cite{xiaomi2025mimo} & 1 (18T) & 1 (4T)  & 2 (2T) & \ding{51} & \ding{51} & \ding{51} &   & \ding{51} & \ding{51} &   & \ding{51} \\
\bottomrule
\end{tabularx}
\end{adjustbox}

\vspace{1mm}
\begin{minipage}{0.95\linewidth}
\scriptsize
\textsuperscript{*} Continued training on Minimax-Text-01. \\
$^\dagger$ Pangu Pro MoE does not have a specific long-context stage. Instead, it is trained with 32K sequence length during the two annealing stages.
\end{minipage}
\end{table*}

\subsection{Mid-training Data Used In Popular Models}
In this section, we review and compare the mid-training datasets adopted in popular models. To facilitate clarity, Table.\ref{fig:data_distribution_in_various_model} provides a structured summary of the data distributions used during mid-training, along with the number of pre-training stages and total training tokens.

The Qwen3 models~\cite{yang2025qwen3} are pre-trained in three stages: first on 30T tokens across 119 languages and dialects, then on 5T high-quality 4K-sequence tokens with increased STEM, coding, reasoning, and synthetic data, and finally on hundreds of billions of tokens for long-context training (75\% 16k–32k, 25\% 4k–16k).

The Phi4 models~\cite{abdin2024phi} are pre-trained on $\sim$10T tokens in two stages: stage 1 with mostly filtered web data, and stage 2 with a mix of synthetic tokens and a smaller portion of ultra-filtered, reasoning-focused web data. For mid training, the models are further trained on 75B curated long-context tokens and 175B recall tokens from earlier stages.

The Llama3 405B models~\cite{Dubey2024TheL3} are pre-trained on 15T multilingual tokens across three stages: 1) initial, 2) long-context (800B tokens), and 3) annealing (final 40M tokens). The data mix comprises 50\% general knowledge, 17\% code, 25\% math/reasoning, and 8\% multilingual tokens, with the annealing stage upsampling the highest-quality sources.

The Apple Foundation Model (AFM)~\cite{gunter2024apple} consists of three stages: (1) pre-training stage on 6.3T tokens with a sequence length of 4k, (2) annealing stage on 1T tokens with a sequence length of 8k, which downweights low-quality webcrawl and upweight code and math data, (3) long context stage on 100B tokens with a sequence length of 32k, using the data mixture from the annealing stage, augmented with synthetic long-context Q\&A data.

The Deepseek-V3 models~\cite{DeepSeekAI2024DeepSeekV3TR} are first pre-trained on 14.8T diverse and high-quality tokens. After that, Deepseek-V3 performs two context extension stages, first trained on 60B tokens with a sequence length of 32K and then trained on 60B tokens with a sequence length of 128K.

The SmolLM2 models~\cite{allal2025smollm2} are trained on 11T tokens across three pre-training stages, one annealing stage, and one context extension stage. Stage 1 uses 6T curated tokens (90\% web, 10\% code; web split into 60\% FineWeb-Edu~\cite{penedo2024fineweb} and 40\% DCLM). Stage 2 trains on 2T tokens (75\% web, 20\% code, 5\% math), and Stage 3 on 2T tokens (74\% web, 16\% code, 10\% math). The annealing stage adds 1T high-quality tokens (58\% web, 24\% code, 14\% math, 4\% synthetic textbooks). Finally, long-context training extends context length from 2k to 8k tokens over 75B tokens, with 40\% long documents and 60\% from the Stage 4 mixture.

The Nemotron-4 models~\cite{adler2024nemotron} are pre-trained on 9T tokens, with 8T for general pretraining and 1T for mid-training. The data mix includes 70\% English, 15\% multilingual, and 15\% code.
In the mid-training phase, two distinct data distributions are employed~\cite{parmar2024nemotron}. The first distribution, which makes up the majority of this phase, consists of tokens already seen during pretraining but reweighted to emphasize higher-quality sources. 
The second distribution introduces a smaller portion QA-style alignment examples and up-weights domains where the model underperforms.

The OLMo 2 models~\cite{olmo20242} adopt a two-stage pre-training process. The first stage uses $\sim$5T tokens of primarily web data. The second mid-training stage employs the Dolmino Mix 1124 dataset, which is smaller but higher-quality and includes synthetic data for strengthening weak capabilities. To maximize generalization, OLMo 2 repeats mid-training with different random orders and averages the resulting models. The 7B model trains three runs of 50B tokens each, while the 13B model performs three runs of 100B tokens plus one of 300B tokens, averaging across all.

The Yi-Lightning model~\cite{wake2024yi} follows a three-stage training paradigm: an initial pre-training stage emphasizing data diversity for broad foundational capabilities; an annealing stage that gradually upsamples high-quality data with emphasis on complex reasoning and low-resource multilingual support; and a fast-decay stage (12.5\% of total tokens) that further strengthens high-quality data usage and introduces early instruction-tuning adaptation. After these three stages, the model undergoes an additional 20B-token training phase to enhance long-context performance.

The Zamba-v1 model~\cite{glorioso2024zamba} is pre-trained in two phases: an initial stage on 950B open web tokens, followed by a mid-training stage with a mixture of 60\% pretraining data and 40\% high-quality datasets. The mid-training set spans over 100 curated sources, including math (StackMathQA), code (EvolInstructCode), instruction tuning (OpenOrca), and synthetic data from stronger models. Most datasets were trained for one epoch, while select high-quality subsets were upsampled for two.

For the MiniMax-M1 model~\cite{chen2025minimax}, training continues from Minimax-Text-01 with an additional 7.5T tokens using optimized mixtures to enhance reasoning and long-context capabilities while preserving diversity. Data quality is improved through refined web/PDF parsing, enhanced cleaning, and semantic de-duplication, prioritizing natural QA pairs over synthetic data. STEM, code, books, and reasoning-related content constitute 70\% of the corpus. Training uses a constant learning rate of 8e-5 for 2.5T tokens, then decays to 8e-6 over 5T. Long-context extension follows a four-stage schedule, expanding the window from 32K to 1M tokens to ensure stability in lightning attention.

The pre-training of Pangu Pro MoE~\cite{tang2025pangu} follows a three-phase process inspired by cognitive development: a general phase (9.6T tokens) for foundational knowledge, a reasoning phase (3T) to strengthen complex reasoning with high-quality STEM, code, and synthetic CoT data, and an annealing phase (0.4T) to refine behavior and transition into instruction tuning. In the reasoning phase, synthetic short- and long-form CoT samples and extended contexts (32K) are introduced to align with long reasoning tasks. The annealing stage increases instruction-style data (20\%) and advanced STEM content (18\%), using curriculum-based sampling and ablation with a 7B proxy to optimize data strategies.

The Hunyuan-Large model~\cite{sun2024hunyuan} is pre-trained on 7T tokens, which contains nearly 1.5T tokens of high-quality and diverse synthetic data. During the learning rate annealing phase, the model is trained on 5\% of the highest-quality pre-training tokens, which plays a pivotal role in augmenting the model's performance. After the learning rate annealing phase, Hunyuan-Large is trained on longer sequences to enable its longer-context capability (up to 256K tokens). The training corpus during long-text phase is consists of 25\% natural long-context data obtained from books and codes and 75\% normal length pre-training data, inspired by~\cite{gao2024train}.

The Hunyuan-A13B model~\cite{yang2025qwen3} is pre-trained on 20T tokens. Then in midt-traininig, it implemented a fast annealing stage on 300B tokens. Following the annealing phase, Hunyuan-A13B progressed through two long-context stages to expand its context length to 32k tokens, then to 256K tokens.

The MiniCPM model~\cite{hu2024minicpm} is trained by a two-stage pre-training strategy. During the general pre-training phase, MiniCPM uses 1T coarse-quality pre-training data, which is abundant and can support continuous training when provided with more computational resources. During the mid-training phase, the model is trained on 20B tokens using a mixture of the pre-training data and high-quality knowledge \& ability-oriented SFT data.

The MAP-Neo model~\cite{Zhang2024MAPNeoHC} is pre-trained in two phases. In the general stage, it uses 4.5T tokens (60\% English, 25\% Chinese, 15\% code) to build general text generation ability. The mid-training phase adds 778B high-quality tokens (74.5\% English, 8.5\% Chinese, 17\% code), with a greater emphasis on instructional and coding data. This enrichment improves robustness and equips the model for complex coding tasks as well as professional domain-specific responses.

The Granite 3.0 models~\cite{granite2024granite} follow a two-stage pre-training setup. In Stage 1, dense and MoE variants are trained on 10T and 8T tokens, respectively, using a mixture of 5\% Web, 11\% Domain, 10\% Code, 10\% Math, 10\% Instruction, 5\% Multilingual, 5\% Academic and 4\% Technical. Stage 2 adds 2T tokens drawn partly from Stage 1 sources, supplemented with high-quality open-source and synthetic corpora under permissive licenses.

MiMo-7B~\cite{xiaomi2025mimo} uses a 3-stage pre-training strategy trained on 25T tokens: Stage 1 (General) focuses on high-quality natural data by removing low-value content and upsampling professional domains. Stage 2 (Annealing) increases mathematics and code data to about 70\% to strengthen specialized skills. Stage 3 (Long-context) incorporates around 10\% synthetic data for tasks like math, code, and creative writing, while extending the context length from 8K to 32K tokens, then to 64K tokens, to improve complex task performance.

\subsection{Insights For Mid-Training Data}
\textbf{Selecting high-quality content improves model performance.}
Unlike early pre-training, which prioritizes scale and coverage, the mid-training phase emphasizes quality. Curated datasets align model representations with reasoning, generalization, and alignment objectives. Ablation studies confirm that filtered corpora consistently outperform unfiltered ones~\cite{Su2024NemotronCCTC, wang2025ultra, penedo2024fineweb}. Different datasets, however, have complementary strengths—for instance, FineWeb-Edu excels on academic benchmarks such as MMLU and ARC, while DCLM performs better on commonsense and reasoning tasks~\cite{penedo2024fineweb, NEURIPS2024_19e4ea30}. Thus, combining diverse high-quality sources is critical for broad performance gains.

\noindent\textbf{Prioritizing educational, coding, and math data boosts performance on reasoning and STEM benchmarks.} SmolLM2~\cite{allal2025smollm2} has conducted extensive ablation studies to refine their mid-training mixtures. SmolLM2 demonstrates consistent gains by replacing low-signal web text with structured educational, math, and coding data, indicating that high-quality, domain-focused corpora provide stronger supervision signals during late-stage training. MiniMax-M1 complements this approach by introducing semantic deduplication and heuristic scoring to prioritize question-answer pairs and STEM-oriented data, thereby enhancing reasoning accuracy and improving long-context generalization. Together, these findings highlight the value of systematically curating high-utility data sources in mid-training to maximize both efficiency and downstream task performance.


\noindent\textbf{The role of instruction-style data during mid-training remains uncertain, with models reporting both benefits and limitations.}
Upsampling high-quality instruction and educational content often improves reasoning and coherence~\cite{Dubey2024TheL3, hu2024minicpm}. LLaMA3~\cite{Dubey2024TheL3} amplifies reliable instruction-following data to boost long-context reasoning without harming general perplexity, while MiniCPM~\cite{hu2024minicpm} shows that introducing such data earlier in mid-training is more effective than deferring it to fine-tuning. However, Deepseek~\cite{bi2024deepseek} reports that adding 5M instruction samples late in pretraining provides gains comparable to SFT, suggesting diminishing returns. This divergence underscores an open question: instruction data can aid reasoning, but its optimal placement in the training pipeline remains unresolved.

\noindent\textbf{Long-context capabilities benefit from careful data source selection, not just sequence length.}
ProLong-8B~\cite{gao2024train} and Phi-4~\cite{abdin2024phi} investigate strategies for long-context optimization. Both find that training solely on long data can hurt performance, and a balanced mixture with high-quality short-context data is critical. Phi-4 shows that naturally long-form content outperforms artificially concatenated data in long-context reasoning. Books and code repositories are highlighted as effective long-context sources.

\noindent\textbf{Domain-specific data is impactful even in small proportions.}
OLMo 2~\cite{olmo20242} finds that even a small fraction of domain-specific data (e.g., math and coding) within the mid-training mix can produce disproportionate performance gains. This underscores the quality-over-quantity principle at this stage, where carefully selected high-utility domains provide strong supervision signals, enhance reasoning ability, and improve transfer to specialized downstream tasks without requiring large-scale rebalancing of the corpus.


\noindent\textbf{Synthetic data substantially enhances training efficiency and generalization.}
The broad usage of synthetic data improves both the quality and diversity of the training corpus, enabling models to acquire richer representations and generalize more effectively to unseen data~\cite{sun2024hunyuan}. Beyond simple data augmentation, synthetic corpora can be tailored to emphasize underrepresented domains, complex reasoning, or structured formats, thereby compensating for limitations in naturally collected text. This makes synthetic data particularly valuable in the mid-training stage, where curated, high-signal samples amplify learning efficiency and reinforce specialized capabilities without requiring large-scale raw corpus expansion.

\noindent\textbf{Reasoning-oriented data is critical for strengthening mathematical and logical capabilities.}
FineMath4+ achieves a 2x improvement on GSM8K and a 6x improvement on MATH compared to InfiMM-WebMath, underscoring the importance of preserving high-quality math corpora with step-by-step reasoning~\cite{allal2025smollm2}. Such data provides explicit demonstrations of problem decomposition and logical inference, allowing models to internalize structured reasoning patterns rather than relying solely on surface heuristics. Incorporating reasoning data during mid-training thus amplifies gains in mathematical problem-solving and enhances generalization to broader reasoning benchmarks.

\noindent\textbf{Moderate repetition and rewriting of high-quality data can substitute for scale.}
OLMo 2~\cite{olmo20242} and related studies explore the role of repetition and rewriting. Repeating high-value data (e.g., math examples) a few times yields performance gains, with diminishing returns beyond a certain point. Additionally, rewriting tasks into more amenable formats (e.g., inline annotations or simplified prompts) dramatically improves performance over structurally similar but unrewritten variants. Charton et al.\ further validate that models trained on smaller repeated datasets can outperform those trained on larger, unrepeated datasets, especially in mathematical reasoning.

\noindent\textbf{Maintaining distributional continuity prevents catastrophic forgetting.}
OpenCoder~\cite{huang2024opencoder} highlights the risk of distributional shift during mid-training, where a mismatch between mid-training and pretraining corpora can lead to catastrophic forgetting and degraded generalization. To address this, OpenCoder retains 84\% of its mid-training data from the original pretraining distribution, ensuring continuity while gradually introducing higher-quality subsets. This strategy stabilizes knowledge retention and preserves broad capabilities, showing the importance of balancing innovation in data selection with consistency in distributional coverage.

\noindent\textbf{Overtraining beyond a critical token budget may degrade downstream tunability.}
A recent study~\cite{springer2025overtrained} on catastrophic overtraining warns that excessive pretraining reduces a model’s adaptability during subsequent fine-tuning, as representations may become overspecialized and less flexible. This highlights the importance of defining careful token budgets and stopping criteria during mid-training to balance knowledge consolidation with downstream tunability.

Collectively, these insights position mid-training not as a mere continuation of pretraining, but as a targeted specialization phase.
It is during mid-training that models are strategically refined, leveraging focused, high-quality, and instruction-rich data to enhance core reasoning, coherence, and long-context understanding. These refinements often lay the groundwork for successful instruction tuning and alignment in downstream stages.

\section{Learning Rate Scheduler}
\label{sec:lrsched}
In this section, we review the literature on learning rate (LR) schedulers. We begin with key factors affected LR scheduling design, then we introduce common types of LR schedulers and the LR schedulers in recent LLMs. We also summarize some valuable key insights for selecting appropriate LR scheduler.

\subsection{Key factors affected LR Scheduling}

{LR schedulers are critical components in optimizing LLMs, as they directly affect optimization stability, convergence speed and generalization performance. 
LR scheduling are largely affected by several factors:
\begin{itemize}[leftmargin=*]
    \item \textit{Warmup setting:} The warmup phase mitigates optimization instability during the early iterations, while the decay phase governs convergence dynamics, typically using cosine decay. Recent analyses \cite{kalra2024warmup} further emphasize the effectiveness of properly setting warmup parameters in stabilizing training and preventing divergence in training Transformer models. Gilmer et al. \cite{gilmer2022loss} empirically show that warm-up reduces loss sharpness and improves optimization conditioning, enabling larger LR.
    \item \textit{Decay strategy:} The decay stage of LR schedulers have also been widely studied. You et al. \cite{you1908does} investigate the role of LR decay in modern neural networks, showing that a large initial LR suppresses the memorization of noisy data, while a gradual decay facilitates the learning of complex patterns.
    \item \textit{Batch size:} Increasing the batch size lowers the variance of stochastic gradient estimates, which in turn supports the use of higher learning rates. Empirical results by Goyal et al.~\cite{goyal2017accurate} demonstrate that, in large mini-batch regimes, employing a cautious initial rate with a properly tuned warm-up phase alleviates optimization difficulties.
\end{itemize}
While these factors provide practical guidance for designing effective schedulers, the broader question of how much LR scheduling truly matters remains debated. Kaplan et al.~\cite{kaplan2020scaling} argued that the final performance of language models is largely insensitive to the specific scheduler, so long as the total LR budget is sufficiently large. In contrast, Hoffmann et al.~\cite{hoffmann2022training} presented empirical evidence that the choice of decay strategy substantially influences optimization dynamics and final model quality, suggesting that scheduler design can have non-trivial consequences in large-scale training.}

\subsection{Common Learning Rate Scheduler Types}

{The design of LR schedules in LLM training has evolved substantially over time. The original Transformer~\cite{vaswani2017attention} introduced a short warm-up phase followed by an inverse square root decay, whose advantage was that it did not require prior knowledge of the total number of training steps. As models and datasets scaled up, however, cosine decay schedules~\cite{loshchilov2016sgdr} became widely adopted in LLM pretraining due to their empirical effectiveness, as seen in GPT-3~\cite{brown2020language}, Gopher~\cite{rae2021scaling}, and Chinchilla~\cite{hoffmann2022training}.}

{Subsequent work questioned whether cosine annealing represents the optimal choice. Defazio et al.~\cite{defazio2023optimal} demonstrated that linear decay consistently outperforms cosine and other schedulers across optimization methods, while Bergsma et al.~\cite{bergsma2025straight} reported similar findings when training LLMs with AdamW. More recently, Ibrahim et al.~\cite{ibrahim2024simple} proposed an “infinite” LR schedule, which eliminates the need for fixed token budgets and repeated warmups by employing a four-phase strategy, thus supporting continual training with reduced forgetting. Despite these advances, many modern LLMs continue to adopt the 10x cosine variant introduced by~\cite{kalra2024warmup}, given its slight empirical advantage over standard cosine decay.}

{Across these developments, most training pipelines converge on a composite structure: a linear warm-up phase, which gradually increases the LR to mitigate instability, followed by a decay phase whose form—cosine, linear, or multi-stage—varies across models. In multi-phase or annealing-style training paradigms, stage-wise schedulers such as WSD apply distinct decay behaviors at different stages to better align with the evolving data distribution and optimization objectives. By modulating the LR in tandem with training phases, these approaches facilitate a smooth transition from rapid exploration in early stages to fine-grained convergence later, thereby enhancing both optimization efficiency and generalization performance.}

\begin{figure}
    \centering
    \includegraphics[width=\linewidth]{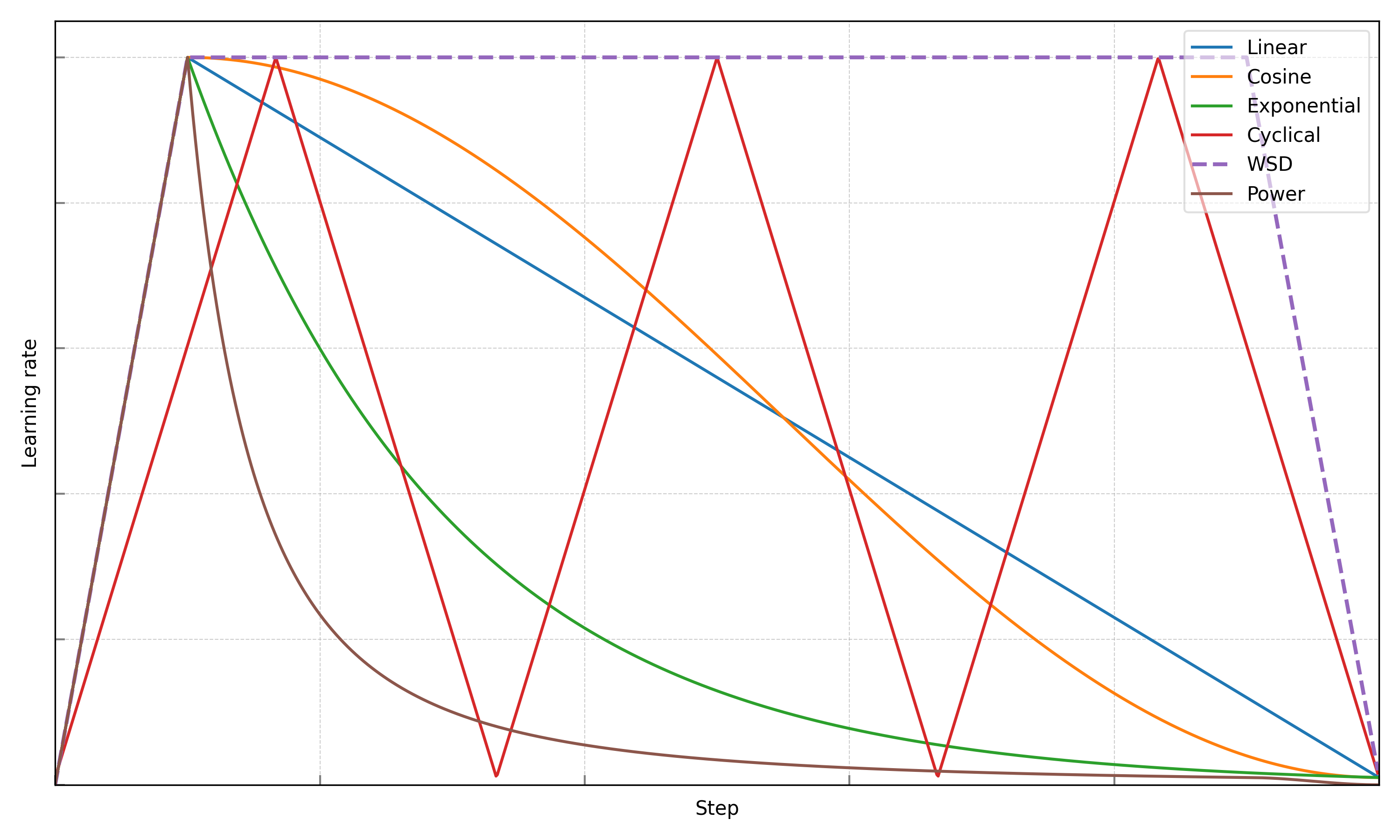}
    \caption{Comparison of different LR schedulers.}
    \label{fig:lrs}
\end{figure}

{In the following, we provide a detailed overview of mainstream decay-phase LR schedulers. Figure~\ref{fig:lrs} provides a unified visualization of commonly used LR schedules (linear, cosine, exponential, cyclical, and WSD), to highlight their characteristic dynamics over training steps. To facilitate a clear comparison, all LR schedulers are set up with the same warmup steps, peak LR, total training steps, and initial and final learning rates of zero.
}

\subsubsection{Linear Scheduler}

The linear LR scheduler was originally proposed in \cite{goyal2017accurate} and has since been widely adopted in large-scale model training.

In the decay phase, the LR decreases linearly from a peak value \(\eta_{peak}\) to a final value \(\eta_{end}\) over \(S_{\text{dc}}\) steps:
\begin{equation}
\eta_t = \eta_{peak} - (\eta_{peak} - \eta_{end}) \cdot \frac{t}{S_{\text{dc}}}, \quad 0 \leq t \leq S_{\text{dc}}.
\end{equation}

This linear decay schedule enables a smooth transition from rapid exploration in early training to more stable convergence in later stages. Its simplicity and effectiveness make it a common choice in both academic and industrial settings.

\subsubsection{Cosine Scheduler}
The cosine LR scheduler typically consists of a linear warmup phase followed by a cosine decay, gradually decreases the LR following a cosine curve after it reaches its maximum after warmup phase.
The decay function is given by
\[
\eta_t = \eta_{end} + \frac{1}{2}(\eta_{peak} - \eta_{end}) \left(1 + \cos\left(\frac{\pi t}{S_{\text{dc}}}\right)\right), \quad 0 \leq t \leq S_{\text{dc}},
\]
where \(\eta_{peak}\) and \(\eta_{end}\) denote the peak and final learning rates, respectively.

A key hyperparameter is the decay length \(S_{\text{dc}}\) at which cosine decay decreases to the minimum for the first time, often set equal to the total number of training steps \(S_{\text{total}}\) when training length is predetermined. Prior studies~\cite{hu2024minicpm, kaplan2020scaling, hoffmann2022training} have shown that both \(S_{\text{dc}} < S_{\text{total}}\) and \(S_{\text{dc}} > S_{\text{total}}\) lead to suboptimal performance. In particular, setting \(S_{\text{dc}} = S_{\text{total}}\) improves training efficiency, as it avoids prematurely decaying the LR or keeping it high for too long. Possible explanation for the effectiveness of cosine LR schedulers with \(S_{\text{dc}} = S_{\text{total}}\) lies in the balance between an extended high LR phase due to cosine decay, which may aid in global exploration, and a full decay phase, which could promote stable convergence dynamics. Consequently, the intermediate checkpoints tend to be suboptimal, complicating the continual pretraining of an existing language model.

\subsubsection{Exponential Scheduler}

The exponential LR scheduler~\cite{li2019exponential} gradually decreases the LR by multiplying it with a fixed decay factor at each step, enabling smooth and continuous reduction throughout training. It is defined as:
\[
\eta_t =  \eta_{init} \cdot e^{-kt}
\]
where \(\eta_{init}\) is the initial LR, and \(k\) is a decay constant that controls the rate of exponential decay.

Its simplicity and continuous decay pattern make it suitable for tasks with stable long-term optimization needs. However, it offers less control and interpretability over scheduling stages and may decay too quickly without careful tuning.

\subsubsection{Knee Scheduler}

The Knee scheduler~\cite{iyer2023wide} is an explore–exploit LR strategy, inspired by the wide-minima density hypothesis, which hypothesizes that narrow minima are significantly higher than wide minima in the loss landscape of deep neural networks.  It comprises two distinct phases: an initial exploration phase, during which the model is trained with a constant high LR for a sufficient duration to increase the likelihood of converging near a wide minimum; and a subsequent exploitation phase, wherein the LR linearly decays to zero without requiring any additional hyperparameters.

\subsubsection{Cyclical Scheduler}

The cyclical scheduler~\cite{smith2017cyclical} cyclically vary between predefined boundary values in a cyclic manner rather than decaying monotonically. The cyclical LR at training step \( t \) is defined as
\begin{align}
\eta_t &= \eta_{init} + (\eta_{peak} - \eta_{init}) \cdot \nonumber \\
&\quad \max\left(0,\ 1 - \left| \frac{t}{s} - 2 \left\lfloor 1 + \frac{t}{2s} \right\rfloor + 1 \right| \right)
\end{align}
where \( \eta_{init} \) and \( \eta_{peak} \) are the minimum and maximum LR, \( s \) is the half-cycle length (i.e., the number of steps from \( \eta_{init} \) to \( \eta_{peak} \)), and \( \lfloor \cdot \rfloor \) denotes the floor function, which returns the greatest integer less than or equal to its input. Its adaptive nature eliminates the need for extensive experimentation to identify optimal learning rates and schedules, while often achieving near-optimal accuracy in fewer iterations.

\subsubsection{WSD Scheduler}
MiniCPM~\cite{hu2024minicpm} proposes Warmup-Stable-Decay (WSD) LR scheduler that divide the training stage into three phases: the warmup stage, the stable training stage, and the decay stage. The function form of WSD is:

\[
WSD(T; s) = 
\begin{cases}
\frac{s}{W} \eta, & s < W \\
\eta, & W \leq s \leq T \\
f(s - T)\eta, & T < s < S
\end{cases}
\]

where $0<f(s-T)\leq1$ is a decreasing function about $s$, $\eta$ is the maximum LR. The WSD scheduler adopts a three-phase structure designed together with two-phase pretraining. It enables efficient convergence by achieving rapid loss reduction in the final decay stage using only 10\% of total tokens, and is particularly effective when high-quality data is introduced mid-training. Compared to cosine schedulers, which require a predetermined total number of steps to achieve optimal decay, WSD achieves comparable performance without this constraint. Its stage-wise design supports resuming from intermediate checkpoints for efficient decay, making it highly flexible for continual pretraining.

\subsubsection{Power Scheduler}

The Power scheduler~\cite{shen2024power} extends from the WSD scheduler and, building on Maximum Update Parameterization ($\mu P$), enables zero-shot LR transfer across different hyperparameters by uncovering a power-law relationship that governs the optimal LR. It is defined as
\begin{equation}
\eta_{\text{power}}(n) = \min\left(\eta_{peak},\; \beta \cdot a n^b\right),
\end{equation}
where $\beta$ is the batch size, $n$ is the number of trained tokens, $a$ and $b$ are the power-law coefficients (amplitude and decay exponent, respectively), and $\eta_{peak}$ is the upper bound on the LR. The final PowerLR scheduler, combined with a warmup and decay stage, is defined as follows:

\begin{equation}
\small
\eta_n = 
\begin{cases}
\frac{n}{N_{\text{}}} \cdot \eta_{\text{power}}(N_{\text{}}) & n < N_{\text{}}, \\
\eta_{\text{power}}(n) & N_{\text{}} \leq n \leq N - N_{\text{decay}}, \\
f(n) \cdot \eta_{\text{power}}(N - N_{\text{decay}}) & n > N - N_{\text{decay}}.
\end{cases}
\label{eq:power}
\end{equation}

The generality of Power scheduler is substantiated by extensive empirical results reported in~\cite{shen2024power}, which confirm its effectiveness across varying hyperparameter scales and model configurations.

\subsubsection{Multi-step Scheduler}
DeepSeek~\cite{bi2024deepseek} replaces the commonly used cosine scheduler with a multi-step LR schedule featuring a linear warmup phase, followed by two discrete decay steps triggered at specific token ratios. Two discrete LR drops are applied at the stage boundaries, enabling efficient continual training by allowing earlier phases to be reused across different training scales.



\subsection{Learning Rate Schedulers in Recent LLMs}

\begin{table*}[ht]
\centering
\caption{Overview of LR Schedulers in representative LLMs.}
\resizebox{\textwidth}{!}{
\begin{tabular}{lllllc}
\toprule
\textbf{Model} & \textbf{LR Decay Scheduler} & \textbf{LR Path} & \textbf{Optimizer} & \textbf{Staged Training} \\
\midrule
GPT-3 (175B)         & Cosine      & (0) $\xrightarrow{\text{}}$ {6e-4} $\rightarrow$ {6e-5} & Adam & \ding{55} \\
Gopher (280B)    & Cosine      & {1e-7} $\xrightarrow{\text{}}$ {4e-5} $\rightarrow$ {4e-6} & Adam    & \ding{55}  \\
Chinchilla (70B)   & Cosine      & {1e-7} $\xrightarrow{\text{}}$ {1.25e-4} $\rightarrow$ {1.25e-5} & AdamW    & \ding{55}  \\
PaLM (540B) & Inverse square root & {1e-2} (constant) $\rightarrow$ {1.98e-5} & Adafactor &  \ding{55} \\
OPT (175B) & Linear (restart\textsuperscript{\dag}) & 0 $\xrightarrow{\text{}}$ {1.2e-4} $\rightarrow$ {1.2e-5} & AdamW & \ding{55} \\
LLaMA (65.2B)  & Cosine       & (0) $\xrightarrow{\text{}}$ {1.5e-4} $\rightarrow$ {1.5e-5} & AdamW    & \ding{55} \\
LLaMA 2 (70B)  & Cosine       & (0) $\xrightarrow{\text{}}$ {1.5e-4} $\rightarrow$ {1.5e-5} & AdamW    & \ding{55}   \\
Jais (13B)  & Linear       & 0 $\xrightarrow{\text{}}$ {1.2e-2} $\rightarrow$ {1.2e-3} & AdamW  &  \ding{55} \\
Qwen (14B)  & Cosine       & 0 $\xrightarrow{\text{}}$ {3e-4} $\rightarrow$ {3e-5} & AdamW    & \ding{55} \\
Falcon (40B)   & Cosine       & (0) $\xrightarrow{\text{}}$ {1.85e-4} $\rightarrow$ {1.85e-5} & AdamW    & \ding{55}   \\
DeepSeek (67B)  & Multi-stage       & (0) $\xrightarrow{\text{}}$ {3.2e-4} $\rightarrow$ {1e-4} $\rightarrow$ {3.2e-5} & AdamW    & \ding{51} \\
OLMo 1.7 (7B)   & Cosine  $\rightarrow$ Linear       & (0) $\xrightarrow{\text{}}$ {3e-4} $\rightarrow$ {3e-5} & AdamW & \ding{51} \\
InternLM2 (20B)  & Cosine & (0) $\xrightarrow{\text{}}$ {3e-4} $\rightarrow$ {3e-5}  & AdamW    & \ding{51} \\
MiniCPM (2.4B)  & WSD & (0) $\xrightarrow{\text{}}$ 0.01 $\rightarrow$ $0.01  \cdot f(s - T)$ & Adam    & \ding{51} \\
DeepSeek V2 (67B)  & Multi-stage & 0 $\xrightarrow{\text{}}$ {2.4e-4} $\rightarrow$ {7.6e-5} $\rightarrow$ {2.4e-5}  & AdamW & \ding{51} \\
Zamba (7B) & Cosine $\rightarrow$ Exponential &  {1.5e-4} $\rightarrow$ {7.5e-5} (cosine); {1.1e-4} $\rightarrow$ {1e-7} (exponential) & Zero-1 distributed &  \ding{51} \\
MAP-Neo (7B) & Two-stage & {2e-5} $\xrightarrow{\text{warmup}}$ {2e-4} $\rightarrow$ {2e-5} (cosine); {2e-4} $\rightarrow$ {1e-4} (exponential) & AdamW & \ding{51}  \\
AFM (3B) & Two-stage & (0) $\xrightarrow{\text{}}$ 0.01 $\rightarrow$ {5e-5} (cosine); (0) $\rightarrow$ {3e-4} $\rightarrow$ {3e-7} & AdamW & \ding{51} \\
LLaMA 3 (405B)   & Cosine $\rightarrow$ Linear & (0) $\xrightarrow{\text{}}$ {8e-5} $\rightarrow$ {8e-7} (cosine) $\rightarrow$ 0 (linear) & AdamW    & \ding{51} \\
Hunyuan-Large (52B out of 389B MoE)   & Three phases & (0) $\xrightarrow{\text{}}$ $\eta_{max}$ $\rightarrow$ $\frac{1}{10}\eta_{max}$ (gradual decay) & AdamW    & \ding{51}   \\
Open-Coder (8B)  & WSD & (0) $\xrightarrow{\text{}}$ {3e-4} $\rightarrow$ {1e-5} & - & \ding{51}  \\
Yi-Lightning   & Three-stage & (0) $\xrightarrow{\text{}}$ $\eta_{\max} \rightarrow 0.5\,\eta_{\max} \rightarrow \eta_{\min}$ & - & \ding{51} \\
Granite 3.0 (8B)  & Power & (0) $\xrightarrow{\text{}}$ 0.02 $\rightarrow$ $4\beta \cdot f(n, N, N_{\text{decay}}) \cdot (N - N_{\text{decay}})^{-0.51}$ & AdamW & \ding{51} \\
Phi-4 (14B)  & Linear & 0 $\xrightarrow{\text{}}$ {3e-4} $\rightarrow$ (0) &  - &  \ding{51}  \\
YuLan-Mini (2.42B)  & WSD & 0 $\xrightarrow{\text{}}$ {1e-2} $\rightarrow$ {5.22e-5} (1-sqrt) &  AdamW & \ding{51}   \\
DeepSeek V3 (37B out of 671B MoE)  & Multi-stage & 0 $\xrightarrow{\text{}}$ {2.2e-4} $\xrightarrow{\text{constant}}$ {2.2e-5} (cosine) $\xrightarrow{\text{constant}}$ {7.3e-6} (constant) &  AdamW & \ding{51}   \\
OLMo 2 (13B)  & Cosine $\rightarrow$ Linear & 0 $\xrightarrow{\text{}}$ {9e-4} $\rightarrow$ {9e-5} (cosine) $\rightarrow$ 0 (linear) & AdamW & \ding{51}  \\
MiniMax-Text-01 (45.9B out of 456B MoE)  & Multi-stage & (0) $\xrightarrow{\text{}}$ {2e-4} $\xrightarrow{\text{constant}}$ {1.3e-4} $\xrightarrow{\text{constant}}$ {3e-5} (exponential) $\rightarrow$ 0 (linear) & AdamW & \ding{51}  \\
SmolLM2 (1.7B)  & WSD & 0 $\xrightarrow{\text{}}$ {5e-4} $\rightarrow$ 0 & AdamW & \ding{51} & \\
xGen-small (9B)  & Revised power~\cite{shen2024power} & - & (Optax lib) & \ding{51}  \\
MiMo (7B) & Multi-stage & 0 $\xrightarrow{\text{}}$ {1.07e-4} $\xrightarrow{\text{constant}}$ {3e-5} (cosine) $\xrightarrow{\text{constant}}$ {1e-5} (cosine) & AdamW & \ding{51}  \\
Pangu Pro MoE (3B)  & Cosine & {3e-4} $\rightarrow$ {1e-7} & AdamW & \ding{51}  \\
MiniCPM4 (8B)  & WSD & N.A. & - & \ding{51}  \\
Hunyuan A13B  & Multi-stage & 0 $\xrightarrow{\text{}}$ {3e-4} $\rightarrow$ {3e-5} (cosine) $\rightarrow$ {8e-6} (rapid cosine) & - & \ding{51}  \\
SmolLM3   & WSD & (0) $\xrightarrow{\text{}}$ {2e-4} $\rightarrow$ 0 & AdamW & \ding{51}  \\
\bottomrule
\end{tabular}
}

\vspace{1mm}  
\noindent
\begin{minipage}{\linewidth}
\textbf{Note:}
The “Staged Training” column indicates distinct pre-training phases prior to supervised fine-tuning, such as data distribution shifts or long-context adaptation. Restart\textsuperscript{\dag} means that LR restarts resume from the checkpointed value. Unless otherwise specified, both the initial and final learning rates are assumed to be zero and denoted as $(0)$. The decay function is defined as \( f(s - T) = 0.5^{\frac{s - S}{5000}} \), where \( S \) denotes the total number of training steps and \( s \) is the current step.
\end{minipage}

\label{tab:lr-schedulers}

\end{table*}

The emergence of LLMs marks a major milestone in the development of NLP. This shift was initiated by the introduction of the Transformer architecture~\cite{vaswani2017attention}, which enabled the development of early large-scale pre-trained models such as BERT~\cite{devlin2019bert} and GPT~\cite{brown2020language}. These early works laid the foundation for scaling both model size and training tokens. The release of GPT-3~\cite{brown2020language} further demonstrated the power of large-scale pre-training, revealing emergent capabilities that arise from the substantial  increase in model parameters and training tokens. Since then, numerous LLMs, including Gopher~\cite{rae2021scaling}, Megatron-Turing~\cite{adler2024nemotron}, Chinchilla~\cite{hoffmann2022training}, PaLM~\cite{chowdhery2023palm}, OPT~\cite{zhang2022opt}, and BLOOM~\cite{workshop2022bloom}, have been released, advancing the frontiers of model scaling, training efficiency, and downstream task performance. In this section, we provide a comprehensive review of the LR schedulers adopted in pre-training phase of the representative LLMs. Furthermore, the mid-training stage plays a critical role in smoothing the transition of learning dynamics and maintaining training stability between pre-training and supervised fine-tuning. We analyze the LR scheduling strategies of LLMs that explicitly incorporate an annealing phase.

With the rapid increase in model size, careful tuning of hyperparameters, including the LR scheduler and batch size, is essential for maintaining training stability and achieving convergence under a limited computational budget. Several works~\cite{mccandlish2018empirical}\cite{kreps2022all}\cite{brown2020language} found that training larger models benefits from the use of larger batch sizes combined with smaller learning rates, contributing to more stable and efficient optimization. The LR scheduler in \textit{GPT-3}~\cite{brown2020language} employs a linear warm-up and followed by a cosine decay, reducing to 10\% of the initial rate, followed by continued training at this reduced level. Consequently, the paradigm shift to focusing heavily on model scaling and pre-training strategies.
\textit{Gopher}~\cite{rae2021scaling} scales transformer-based models up to 280 billion parameters, achieving strong performance across a range of benchmarks. It utilizes the Adam optimizer with a warm-up that increases the LR from $10^{-7}$ to a peak value, followed by cosine decay by a factor of 10. 
Building on this, \textit{Chinchilla}~\cite{hoffmann2022training} investigates the optimal scaling between model size and training tokens under a fixed FLOPs budget, concluding that model size and tokens should scale proportionally. The hypothesis that aligning the cosine decay schedule length with the total number of training tokens improves performance is empirically validated, and is thus adopted in \textit{Chinchilla}. Besides, they empirically show that AdamW outperforms Adam in large-scale training, and therefore adopt AdamW as the optimizer for \textit{Chinchilla}.
PaLM~\cite{chowdhery2023palm} adopts Adafactor~\cite{shazeer2018adafactor} with a fixed LR in the early phase, followed by an inverse square root decay with respect to the training step, and incorporates dynamic weight decay along with empirically tuned hyperparameters to ensure stability in large-scale training.
\textit{LLaMA}~\cite{touvron2023llama, touvron2023llama2, dubey2024llama} form a family of open-weight transformer-based language models developed by Meta, with parameter scales ranging from 7B to 405B across versions. LLaMA 1 and 2 use the AdamW optimizer and a LR schedule consisting of a linear warm-up followed by cosine decay to 10\% of the peak LR. LLaMA 3 extends this recipe into a three-stage process, using a longer linear warm-up and cosine decay to 1\% of the peak LR during the initial pre-training phase, followed by a final annealing phase that linearly decays the LR to zero over the final 40 million tokens. 
Similarly, Qwen~\cite{bai2023qwen} uses a cosine LR schedule with a specified peak LR followed by a decay to 10\% of the peak LR. In Qwen 2.5~\cite{qwen2025qwen25technicalreport}, the authors further explore scaling laws to optimize hyperparameters across both dense and Mixture-of-Experts (MoE) models, enabling efficient training and performance parity. The pretraining process is restructured into three stages, general stage, reasoning stage and long context stage, with accelerated LR decay during the reasoning stage to facilitate more effective optimization in Qwen 3~\cite{yang2025qwen3}.
\textit{Nemotron-4}~\cite{parmar2024nemotron, adler2024nemotron} introduces an additional continued training phase, employing two distinct data distributions and a steeper LR decay slope over absolute magnitude to help the model transition smoothly from pre-training corpus and better learn newly emphasized corpus.
\textit{DeepSeek}~\cite{bi2024deepseek} and \textit{DeepSeek-V2}~\cite{liu2024deepseek} employ a warmup-and-step-decay LR schedule, consisting of a 2K-step linear warm-up followed by discrete drops at fixed training milestones, a design that facilitates continual training reuse while achieving performance comparable to that of cosine decay. \textit{DeepSeek-V3}~\cite{liu2024deepseekv3} extends this scheduling strategy by introducing a plateau phase followed by cosine decay after the warm-up, and subsequently transitions into a long-context extension stage.
Meanwhile, \cite{hu2024minicpm} put forward the Warmup-Stable-Decay (WSD) LR scheduler, which closely resembles that of \textit{DeepSeek} and is likewise tailored for continual training, enabling effective reuse of intermediate model checkpoints. Their experiments show that, during the decay stage, the loss rapidly drops as the LR decreases, reaching or even falling below that of the Cosine schedule at step $T = S$. 
The Power scheduler~\cite{shen2024power}, developed through further research based on WSD and $\mu$P, introduces a new LR schedule that combines a linear warmup phase, a slow power-law decay, and a fast exponential decay. Granite 3.0~\cite{granite2024granite} adopts the Power scheduler to train its lightweight foundation models. 

Beyond the mainstream LLMs discussed above, a variety of LR schedulers have been explored in LLM. Among them, the \textit{cosine LR scheduler} is the most widely adopted~\cite{kaplan2020scaling, hoffmann2022training, rae2021scaling, gunter2024apple, cai2024internlm2, touvron2023llama, touvron2023llama2, dubey2024llama, brown2020language, olmo20242, tang2025pangu, bai2023qwen, almazrouei2023falcon}, where the LR increases linearly during the warm-up phase and then decays following a cosine curve.
Another commonly used approach is the linear LR scheduler, where the LR increases linearly during warm-up and then decreases linearly to zero or a minimal value over the which linearly increases the LR during warm-up and then decays it linearly to zero or a small constant, is another widely adopted strategy~\cite{zhang2022opt, abdin2024phi, bergsma2025straight, defazio2023optimal}.
\cite{glorioso2024zamba, ibrahim2024simple, hu2024minicpm} find it better to rewarm the LR followed by a rapid exponential (instead of linear) decay.
However, some studies suggest that linear decay may not always be optimal. \cite{glorioso2024zamba, ibrahim2024simple, hu2024minicpm} argue that exponential decay leads to better performance than linear decay.
In addition, the inverse square root LR scheduler, typically combined with the Adafactor optimizer, has been adopted by several models, including \textit{T5}~\cite{raffel2020exploring}, \textit{PaLM}~\cite{chowdhery2023palm}, and \textit{OpenMoE}~\cite{xue2024openmoe}.

In addition to those continuous decay strategies, some models have adopted variants of the multi-step LR scheduler. \textit{DeepSeek} employs discrete LR drops at fixed milestones, while \textit{MiniCPM} integrates the WSD scheduler; both strategies facilitates checkpoint reuse and support continual training. Empirical results from both models show that the multi-step LR scheduler performs comparably to cosine decay during pretraining. With the introduction of the annealing stage, the WSD strategy has been adopted in several recent works~\cite{hu2024yulan, allal2025smollm2, huang2024opencoder}, owing to its flexibility in accommodating multi-stage pretraining.
\textit{MAP-Neo}, extending from \textit{MiniCPM}, employs a two-stage scheduler combining warm-up and cosine decay, followed by a subsequent exponential decay phase. Similarly, an increasing number of models~\cite{sun2024hunyuan,li2025minimax,wake2024yi,nijkamp2025xgen}, incorporate multiple pretraining stages, typically involving distribution shifts and/or sequence length extensions, and accordingly adopt multi-phase LR strategies with distinct designs. A recent trend emphasizes the use of an explicit annealing phase, which has shown effectiveness in improving optimization. Models that explicitly adopt such an annealing stage~\cite{} typically employ multi-stage LR schedules, such as those used in \cite{hu2024minicpm} and \cite{bi2024deepseek}.
Table~\ref{tab:lr-schedulers} provides a comparative summary of LR schedulers and optimizer settings, as explicitly reported for representative LLMs in official sources.

\subsection{Key Insights}

\noindent\textbf{Linear warm-up with cosine/linear decay is widely used but largely heuristic.}
This scheme is the most common choice in LLM training due to its empirical stability, even though its design is guided more by practice than theory. While several alternatives, such as the infinite LR schedule~\cite{ibrahim2024simple}, have been proposed to relax assumptions like fixed token budgets and repeated warm-ups, they often lack robustness across architectures and tasks. Consequently, cosine and linear decay remain the prevailing strategies, reflecting both their demonstrated empirical reliability and the broader reliance on heuristic tuning in the field.

\noindent\textbf{Warm-up helps stability.}
Warm-up phases are known to alleviate gradient instability and loss sharpness in early training~\cite{goyal2017accurate, gilmer2022loss, kalra2024warmup}, thereby facilitating the use of larger peak learning rates. However, the optimal duration and scaling of warm-up still lack theoretical justification and vary significantly across model scales and optimizers.

\noindent\textbf{Decay strategies remain unsettled and system-dependent.}
The design of the decay phase in LR schedules is far from resolved. Although cosine annealing has been widely adopted in LLM training~\cite{brown2020language, rae2021scaling, hoffmann2022training}, recent empirical analyses suggest that linear decay can yield superior convergence and generalization in certain settings~\cite{defazio2023optimal, bergsma2025straight}. Beyond these comparisons, the effectiveness of a decay strategy interacts intricately with factors such as batch size, model scale, and optimizer dynamics, leading to system-dependent behaviors that are not yet theoretically characterized.

\noindent\textbf{Scaling laws help LR selection.}
Empirical scaling laws provide practical guidance for choosing learning rates across model sizes and training regimes, though they remain incomplete and sensitive to model-specific factors. Kaplan et al.~\cite{kaplan2020scaling} uses scaling law to show that the optimal LR depends on the target loss: smaller values are needed near convergence for stability, whereas larger rates can be effective in short, compute-limited runs. Their analysis also indicated that larger models require smaller learning rates, while smaller models tolerate more aggressive ones. More recently, Li et al.~\cite{li2025predictable} proposed a universal scaling law for hyperparameter selection in LLM pretraining, finding that a fixed final LR leads to more stable convergence.

\noindent\textbf{Current evidence highlights both the critical role and the unresolved uncertainties of LR scheduling.}
Conflicting empirical results, for example, the claim that final performance is insensitive to schedule shape under sufficient LR budget~\cite{kaplan2020scaling}, versus evidence of significant variance across decay types~\cite{hoffmann2022training, defazio2023optimal}, highlight the lack of consensus and reproducible benchmarks in this space.
\section{Long Context Extension}
\label{sec:longcontext}
In this section, we focus on the long context extension phases in popular LLMs.

\subsection{Background and Formulation}
\subsubsection{Self‑attention with Positional Encoding}
\label{sec:prelim-sa}

We briefly fix notation and recall scaled dot–product self‑attention.
Let $\mathbb S_N=\{w_i\}_{i=1}^{N}$ be an $N$‑token sequence and
$\mathbb E_N=\{\mathbf x_i\}_{i=1}^{N}$ the corresponding token embeddings with
$\mathbf x_i\in\mathbb R^{d}$.
To inject order information, we use task‑specific maps
$f_q,f_k,f_v$ that take the \emph{content} embedding $\mathbf x_i$ together with its
\emph{position} index $i$ and produce the query, key and value vectors:
\begin{equation}
\label{eq:qkv}
\mathbf q_{m}=f_q(\mathbf x_{m},m),\quad
\mathbf k_{n}=f_k(\mathbf x_{n},n),\quad
\mathbf v_{n}=f_v(\mathbf x_{n},n).
\end{equation}
Given \eqref{eq:qkv}, single‑head scaled dot–product attention from position $m$ to all
positions $n\in\{1,\dots,N\}$ computes
\begin{equation}
\label{eq:attn-weights}
a_{m,n}\;=\;
\frac{\exp\!\bigl(\mathbf q_{m}^{\top}\mathbf k_{n}/\sqrt d\bigr)}
     {\displaystyle\sum_{j=1}^{N}\exp\!\bigl(\mathbf q_{m}^{\top}\mathbf k_{j}/\sqrt d\bigr)},
\qquad
\mathbf o_{m}\;=\;\sum_{n=1}^{N}a_{m,n}\,\mathbf v_{n},
\end{equation}
where $\mathbf o_m$ denotes the output at token $w_m$.
For clarity, we omit the (standard) causal mask and multi‑head split; all subsequent
derivations apply per head identically.

\subsubsection{Rotary Position Embedding (RoPE)}
\label{sec:rope}

RoPE encodes positions by \emph{rotating} the query/key vectors in a set of
two‑dimensional sub‑planes of the embedding space.
Let $W_q,W_k\in\mathbb R^{d\times d}$ denote the linear maps that first produce the
\emph{base} query/key vectors
$\tilde{\mathbf q}_m=W_q\mathbf x_m$ and $\tilde{\mathbf k}_n=W_k\mathbf x_n$.
A position‑dependent rotation is applied after the projections:
\begin{align}\label{eq:rope-rotate-qk}
\mathbf q_m
    = f_q(\mathbf x_{m},m)
    = R^d_{\Theta,m}\,\tilde{\mathbf q}_m
    = R^d_{\Theta,m}\,W_q\mathbf x_m, 
\\
\mathbf k_n
    = f_k(\mathbf x_{n},n)
    = R^d_{\Theta,n}\,\tilde{\mathbf k}_n
    = R^d_{\Theta,n}\,W_k\mathbf x_n .
\end{align}
Here $R^d_{\Theta,m}$ is an orthogonal block‑diagonal rotation that acts independently
on $d/2$ disjoint 2‑D sub‑planes:
\begin{equation}
\label{eq:R-block}
R^d_{\Theta,m}
=\bigoplus_{j=1}^{d/2}
\begin{pmatrix}
\cos(m\theta_j) & -\sin(m\theta_j)\\
\sin(m\theta_j) &  \cos(m\theta_j)
\end{pmatrix},
\end{equation}
where $\bigoplus$ concatenates the $2{\times}2$ blocks along the embedding dimension,
$\Theta=\{\theta_{j}\mid j=1,2,\dots,d/2\}$ is the set of base angular frequencies, and
each block rotates the coordinate pair
$\bigl(\tilde q_{m,2j-1},\tilde q_{m,2j}\bigr)$ (and analogously for
$\tilde{\mathbf k}_n$).
The original RoPE schedule sets
\[
\theta_j = b^{-\frac{2(j-1)}{d}}, \qquad b=10000, \qquad j=1,\dots,d/2 .
\]
Intuitively, the $j$‑th 2‑D sub‑plane behaves like a complex plane whose phase advances
linearly with the position index $m$ at rate $\theta_j$.

A crucial property of RoPE is that inner products of rotated vectors depend only on
\emph{relative} positions. Because $R^d_{\Theta,m}$ is orthogonal and
${R^d_{\Theta,m}}^{\top} R^d_{\Theta,n}=R^d_{\Theta,{m-n}}$, 
the attention dot‑product becomes
\begin{equation}\label{eq:rope-relative}
\mathbf q_m^{\top}\mathbf k_n
  = \tilde{\mathbf q}_m^{\top} R^d_{\Theta,{m-n}} \tilde{\mathbf k}_n
  = \mathbf x_{m}^{\top} W_q^{\top} R^d_{\Theta,{m-n}} W_k\,\mathbf x_n ,
\end{equation}
which depends only on the displacement $m-n$.
This relative‑position invariance is at the heart of many long‑context strategies:
with suitable frequency remapping, models trained at a certain window length can
generalize to much longer contexts during fine‑tuning or inference.

\subsection{A Unified Overview of Context Extension Methods}
\label{sec:rope-h-only}

\begin{table*}[t]
\centering
\small
\caption{Unified frequency-scaling view of long-context families (fix absolute index $g(m)=m$). Let $s=L_{\text{new}}/L_{\text{train}}$. All methods differ only in how they remap RoPE base frequencies $\theta \mapsto h(\theta)$; attention uses $\cos(m\,h(\theta)),\sin(m\,h(\theta))$.}
\label{tab:unified-h-theta}
\resizebox{\textwidth}{!}{%
\begin{tabular}{p{2cm} p{4.6cm} p{2.7cm} p{4.8cm} p{4.8cm} p{4.8cm}}
\toprule
\textbf{Family} & \textbf{Unified map $h(\theta)$} & \textbf{Extra stabilizer} & \textbf{Fixes vs previous} & \textbf{Pros} & \textbf{Cons} \\
\midrule
PI (Positional Interpolation)
& $h(\theta)=\theta/s$ \, (uniform) 
& None
& Avoids naive extrapolation by compressing positions into the seen range
& Conceptually simple; light tuning; robust up to moderate $s$
& Uniform compression hurts high-frequency (local) signals; short-context quality may drop at large $s$ \\
\addlinespace
NTK-aware (ABF-style)
& $h(\theta)=\theta/s^{\rho(\theta)}$, with $\rho(\theta)\!\in[0,1]$ increasing toward low frequencies
& None
& Reduces long-range attention decay vs.\ PI while keeping more locality
& Better locality/short-length retention than PI; works with minimal tuning
& Hyperparameter coupling (e.g., target length vs.\ curvature); mid-band still mixes interp/extrap \\
\addlinespace
NTK-by-part (piecewise)
& $\displaystyle
h(\theta)=\begin{cases}
\theta & \text{(high freq)}\\[2pt]
\theta/s & \text{(low freq)}\\[2pt]
\theta/s^{\rho(\theta)} & \text{(mid freq)}
\end{cases}$
& None
& Preserves the shortest wavelengths (local detail) while still extending range
& Strong short-context preservation; cleaner trade-off across bands
& More boundaries/knobs to set; modest tuning often needed \\
\addlinespace
YaRN
& Same piecewise $h(\theta)$ as NTK-by-part
& Length-dependent logit temperature / scaling $t(s)$
& Stabilizes loss/perplexity for fast extension; reduces data/steps vs.\ prior methods
& Efficient route to 32k–128k; robust convergence with few steps
& Extra temperature and schedule to choose; still handcrafted frequency policy \\
\addlinespace
LongRoPE
& Non-uniform $h(\theta)$ discovered by search; applied progressively (stages)
& Progressive extension + short-length readjustment
& Surpasses handcrafted rules by optimizing $h$; avoids catastrophic angles at ultra-long lengths
& Reaches $\sim$\,2M context with limited extra steps; recovers short-length performance
& Pipeline complexity (search + stages + readjustment); fewer closed-form guarantees \\
\bottomrule
\end{tabular}%
}
\vspace{-4pt}
\end{table*}

\noindent\textbf{Goal.} We present a \emph{single} formulation that covers five representative long-context families---\textbf{PI}, \textbf{NTK}, \textbf{NTK-by-part}, \textbf{YaRN}, and \textbf{LongRoPE}---while keeping the \emph{absolute position} fixed. This isolates what changes in all methods: the \emph{frequency spectrum} used by RoPE.

\noindent\textbf{Unified formulation for context extension.}
We introduce a single scalar function
\[
  h:\theta_j \;\mapsto\; h(\theta_j)
\]
that rescales each base frequency $\theta_j$.  
Any of these extended‑context position embedding $f'$ can be expressed with the original rope embedding $f$ as 
\begin{equation}
  f^{\prime}_W(\mathbf x_m,m,\theta_j)
  \;=\;
  f_W\!\bigl(\mathbf x_m,\,m,\,h(\theta_j)\bigr)
  \;=\;
  \mathbf R_{h(\theta_j),m}\,W\,\mathbf x_m, 
  \label{eq:rope-extrapol-h}
\end{equation}
where $W\in \{W_q, W_k\}$. These methods are summarize in Table~\ref{tab:unified-h-theta}.




\begin{figure}
    \centering
    \includegraphics[width=1\linewidth]{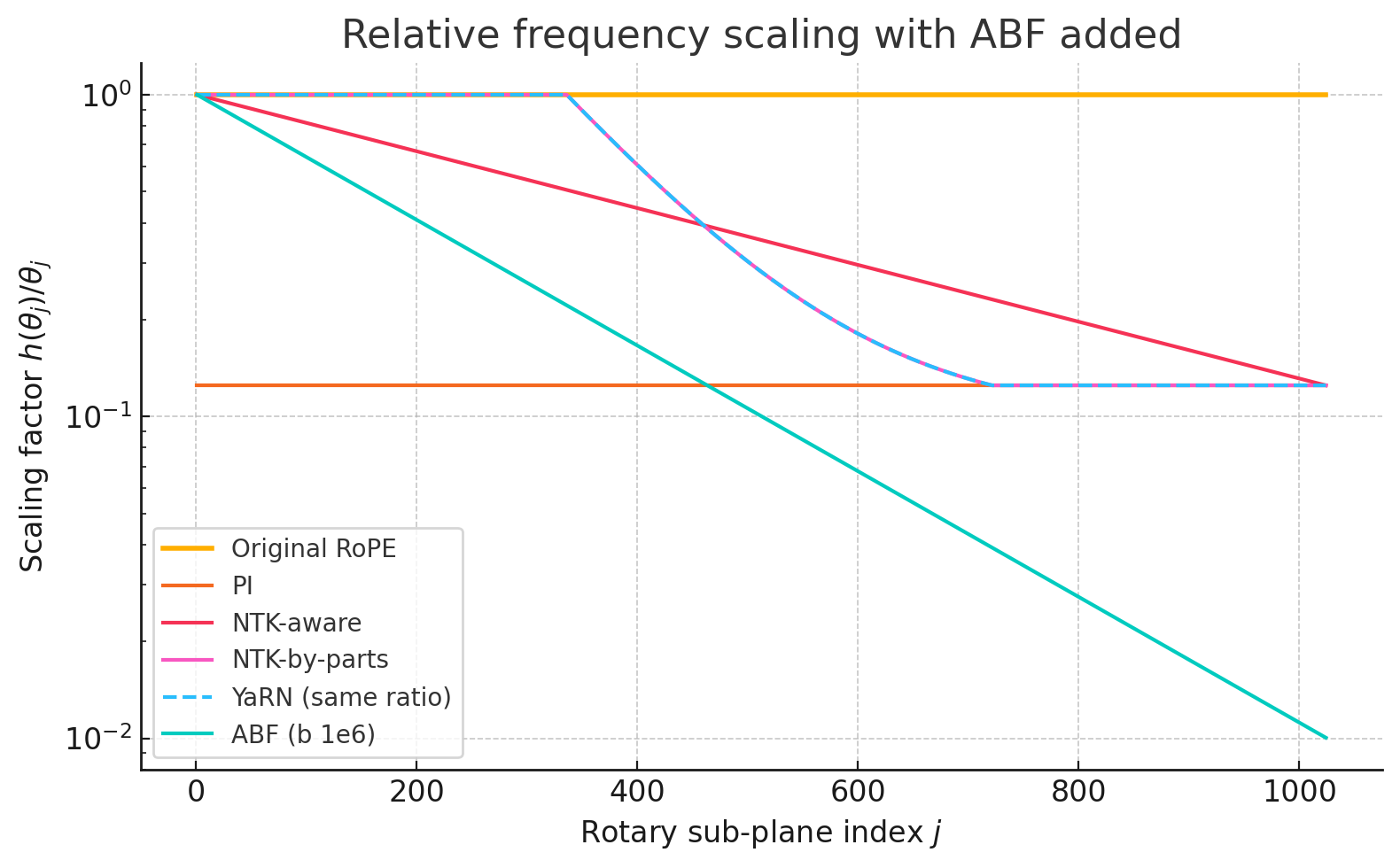}
    \caption{Comparison of frequency remapping schemes.}
    \label{fig:scaling-ratio}
\end{figure}

As shown in Fig.~\ref{fig:scaling-ratio}, all concrete methods now differ \emph{only} by their choice of
$h(\theta_j)$ as a (possibly $j$‑dependent) function of $s$.

\subsubsection{Position Interpolation (PI)}
Position Interpolation rescales absolute positions when evaluating sequences longer than the pre‑training window~\cite{chen2023extending}.  Let $L_{\text{train}}$ and $L_{\text{target}}>L_{\text{train}}$ be the maximum lengths during training and inference. 

Each position $m$ is mapped to
$
m'=\alpha m,\qquad \alpha=(L_{\text{train}}-1)/(L_{\text{target}}-1),
\label{eq:pi_scale}
$ and $h_{\text{PI}}(\theta_j),m$ is defined as 
$$h_{\text{PI}}(\theta_j)=\alpha \theta_j.$$ 
\begin{definition}[Position Interpolation]\label{def:pi}
\begin{equation}
  f_W^{\text{PI}}(\mathbf x_m,m,\theta_j)
  \;=\;
  \mathbf R_{h_{\text{PI}}(\theta_j),m}\,
  W\,\mathbf x_m 
  \label{eq:pi-final}
\end{equation}
 This equals to setting the rotary angle $h(\theta_j)$ to $\alpha \theta_j$ in Eq.~\eqref{eq:rope-extrapol-h}.  This preserves the overall frequency range while re‑using trained low‑frequency dimensions. PI is easy to implement, but it compresses frequency bands uniformly and might affect the model performance in short problems.
\end{definition}

\subsubsection{Neural Tangent Kernel (NTK) -aware Interpolation}
Inspired by NTK theory~\cite{tancik2020fourier}, the neural networks have trouble learning high‑frequency information when the input dimension is low. This might be the reason why the perplexity of PI deteriorates a bit.

To solve the problem of losing high-frequency information, the NTK-aware~\cite{bloc97_ntk_rope_2023} interpolation is developed. The idea is to extend the low-frequency dimensions more and the high-frequency dimensions less.

With the scale factor
$
  s=L_{\text{target}}/L_{\text{train}}\;>1,
$
define
$
  b' \;=\; b \, s^{d/(d-2)},
  \label{eq:ntk-base}
$
and set the frequency‑remapping function to
\begin{equation}
  h_{\text{NTK}}(\theta_j)
  \;=\;
  (b')^{-\tfrac{2j}{d}},
  \qquad
  j=0,\dots,d/2 .
  \label{eq:ntk-h}
\end{equation}

\begin{definition}[NTK‑aware interpolation]\label{def:ntk-aware}
Using $g(m)=m$ and the map $h_{\text{NTK}}$ in
Eq.~\eqref{eq:rope-extrapol-h} yields
\begin{equation}
  f_W^{\text{NTK}}(\mathbf x_m,m,\theta_j)
  \;=\;
  \mathbf R_{h_{\text{NTK}}(\theta_j),m}\,
  W\,\mathbf x_m .
  \label{eq:ntk-final}
\end{equation}
\end{definition}

Equation~\eqref{eq:ntk-h} spreads the interpolation pressure across
dimensions as shown in Figure~\ref{fig:scaling-ratio}: low‑frequency dimensions (small $j$) receive a larger
scaling $h(\theta_j)/\theta_j$, while high‑frequency ones are
affected less, thereby preserving fine‑grained relative information
and reducing perplexity degradation at long context lengths. But it still blends interpolation/extrapolation in mid frequencies and affects performance under short context.

\subsubsection{NTK‑by‑parts Interpolation}
\label{sec:ntk-by-parts}

Recall the wavelength of the $j$‑th rotary pair:
\[
  \lambda_j = 2\pi\,b^{\tfrac{2(j-1)}{d}},
  \qquad j = 1,\dots,\tfrac{d}{2}.
\]

Given a context size, some dimensions $j$ have a wavelength $\lambda_j$ longer than the maximum context length seen during the pretraining, so these dimensions' embeddings might not be trained on some part of the rotational domains. In such cases, these dimensions all have unique position pairs, making them behave like absolute position embeddings. 

For other dimensions, all tokens become closer to each other, so the dot product of two vectors rotated by a lesser amount is bigger. This has a negative impact on the LLM's ability to understand small local relationships between nearby embeddings. 

So for dimensions with wavelength smaller than context size, we do not interpolate; for wavelengths bigger than the context size, we only interpolate and avoid extrapolations, unlike the NTK-aware method; for dimensions in-between, we extrapolate like the NTK-ware method. 

\noindent\textbf{Relative ratio:}
Let $L$ be the pre‑training context length.  
Define the dimension‑specific ratio
\begin{equation}
  r(j)
  \;=\;
  \frac{L}{\lambda_j}
  \;=\;
  \frac{L}{2\pi\,b^{\tfrac{2(j-1)}{d}}}.
  \label{eq:ratio-r}
\end{equation}
Intuitively, $r(j)$ measures how many full rotary periods fit inside
the training window for dimension~$j$.

\noindent\textbf{Ramp (mask) function:}
Choose two hyper‑parameters $\alpha<\beta$ that delimit three
frequency regimes and define
\begin{equation}
  \gamma\!\bigl(r(j)\bigr)
  \;=\;
  \begin{cases}
    0, & r(j)<\alpha,\\[6pt]
    1, & r(j)>\beta,\\[6pt]
    \dfrac{r(j)-\alpha}{\beta-\alpha}, & \text{otherwise}.
  \end{cases}
  \label{eq:ramp}
\end{equation}
Thus $\gamma=0$ flags \emph{low‑frequency} dimensions
($\lambda_j\!>\!L/\alpha$), $\gamma=1$ flags
\emph{high‑frequency} dimensions
($\lambda_j\!<\!L/\beta$), and the linear middle section creates a
smooth transition.

\noindent\textbf{Frequency map:}
With scale factor $s\!=\!L_{\text{target}}/L_{\text{train}}>1$, set
\begin{equation}
  h_{\text{NTP}}(\theta_j)
  \;=\;
  \bigl(1-\gamma(r(j))\bigr)\,\frac{\theta_j}{s}
  \;+\;
  \gamma(r(j))\,\theta_j,
  \label{eq:h-ntp}
\end{equation}
which \emph{interpolates} low‑frequency dimensions by the factor
$1/s$ (PI‑like) while \emph{leaving} high‑frequency dimensions
unchanged; mid‑range dimensions receive a weighted mix, as shown in Figure~\ref{fig:scaling-ratio}. 

\begin{definition}[NTK‑by‑parts interpolation]\label{def:ntk-by-parts}
Using $g(m)=m$ and the map $h_{\text{NTP}}$ from
Eq.~\eqref{eq:h-ntp} in the generic template
Eq.~\eqref{eq:rope-extrapol-h} yields
\begin{equation}
  f_W^{\text{NTP}}(\mathbf x_m,m,\theta_j)
  \;=\;
  \mathbf R_{\,m\,h_{\text{NTP}}(\theta_j)}\,W\,\mathbf x_m .
  \label{eq:ntp-final}
\end{equation}
\end{definition}

Recommended defaults for LLaMA‑family models are
$\alpha=1$ and $\beta=32$; these can be tuned per architecture
and target length.

\subsubsection{YaRN}
\label{sec:yarn}

YaRN augments the \textit{NTK‑by‑parts} frequency map
$h_{\text{NTP}}(\theta_j)$ (Sec.~\ref{sec:ntk-by-parts}) with a
\emph{temperature} $t>0$ applied to the attention logits:
\begin{equation}
  \operatorname{softmax}\!\Bigl(
     \tfrac{\mathbf q_m^{\top}\mathbf k_n}{\,t\sqrt{d}\,}
  \Bigr),
  \label{eq:yarn-softmax}
\end{equation}
where $d$ is the model dimension.
Empirically, the following scale‑dependent rule provides near‑optimal
perplexity across LLaMA‑family models:
\begin{equation}
  \sqrt{\smash[b]{\tfrac1t}}
  \;=\;
  \alpha(s)
  \;=\;
  0.1\,\ln s + 1,
  \qquad
  s=\frac{L_{\text{target}}}{L_{\text{train}}}>1 .
  \label{eq:yarn-alpha}
\end{equation}
Hence $t=\alpha(s)^{-2}$.

\noindent\textbf{Implementation via length scaling:}
Because RoPE can be viewed as a bank of $2\times2$ rotation matrices,
Eq.~\eqref{eq:yarn-softmax} is equivalent to scaling the \emph{RoPE
embeddings themselves}:
\[
  \mathbf q_m^{\text{YaRN}}
    \;=\;
    \alpha(s)\,
    \mathbf q_m^{\text{NTP}},
  \qquad
  \mathbf k_n^{\text{YaRN}}
    \;=\;
    \alpha(s)\,
    \mathbf k_n^{\text{NTP}},
\]
where $\mathbf q^{\text{NTP}}$ and $\mathbf k^{\text{NTP}}$ use the
frequency remapping $h_{\text{NTP}}(\theta_j)$ from
Eq.~\eqref{eq:h-ntp}.  
No changes to the attention kernel are required; RoPE embeddings are
scaled once and cached for all forward passes.

\begin{definition}[YaRN]\label{def:yarn}
Combine
\[
  g(m)=m,\quad
  h(\theta_j)=h_{\text{NTP}}(\theta_j),\quad
  \text{and}\quad
  \alpha(s)\text{ from Eq.~\eqref{eq:yarn-alpha}}
\]
to obtain
\begin{equation}
  f_W^{\text{YaRN}}(\mathbf x_m,m,\theta_j)
  \;=\;
  \alpha(s)\,
  \mathbf R_{\,m\,h_{\text{NTP}}(\theta_j)}\,
  W\,\mathbf x_m.
  \label{eq:yarn-final}
\end{equation}
\end{definition}

Recommended values for Eq.~\eqref{eq:yarn-alpha} were obtained by
fitting $\alpha(s)$ on LLaMA‑7B/13B/65B for scale factors
$s\!\in\![1,512]$ without fine‑tuning; the same rule transfers well
to Llama‑2 models (7B, 13B, 70B).\footnote{%
  For alternative kernels such as Flash‑Attention 2 the same
  embedding‑side scaling applies because the softmax temperature in
  Eq.~\eqref{eq:yarn-softmax} is absorbed into $\alpha(s)$.} 
YaRN proves itself in many models, but it still requires extra hyperparameter tuning for $\alpha(s)$. 

\subsubsection{LongRoPE}
\label{sec:longrope}

Unlike PI, NTK-aware or YaRN—which apply a \emph{global}
dimension-wise rule—\textbf{LongRoPE}~\cite{ding2024longrope}  learns a \emph{token-dependent}
scale for every rotary sub-plane.  
Let \(P=\{p_0,p_1,\dots,p_K\}\) be a set of \emph{anchor} positions
with \(0=p_0<\dots<p_K=L_{\text{target}}\).
For each dimension \(j\in[1,\tfrac{d}{2}]\) and every interval
\([p_k,p_{k+1})\) we introduce a learnable scale
\(\sigma_{j,k}>0\).
The frequency map therefore becomes a piece-wise constant function
\begin{equation}
  h_{\text{LR}}\bigl(\theta_j,m\bigr)
  \;=\;
  \sigma_{j,k}\,\theta_j,
  \qquad
  p_k\le m < p_{k+1}.
  \label{eq:h-longrope}
\end{equation}

\noindent\textbf{Optimisation objective:}
Given the pretrained attention logits
\(z_{m,n}=\mathbf q_m^{\top}\mathbf k_n\)
(\(L_{\text{train}}\!\times\!L_{\text{train}}\) region),
LongRoPE selects \(\{\sigma_{j,k}\}\) by minimising the discrepancy
between the original logits and those produced with the scaled RoPE:
\begin{equation}
  \min_{\{\sigma_{j,k}\}}
  \; \mathbb{E}_{(m,n)\sim\mathcal D}
  \bigl\|
    z_{m,n}
    -\tilde z_{m,n}
  \bigr\|_2^2,
  \label{eq:longrope-loss}
\end{equation}
where \(\tilde z_{m,n}\) is obtained from
Eq.~\eqref{eq:rope-extrapol-h} with
\(h(\theta_j)=h_{\text{LR}}(\theta_j,m)\).
The sampling distribution \(\mathcal D\) mixes pairs
\((m,n)\) \emph{within} pre-training window
(\(m,n<L_{\text{train}}\)) and \emph{out-of-window}
pairs (\(m\) or \(n\ge L_{\text{train}}\)); this encourages model
to preserve short-range behaviour while smoothly extending to
\(L_{\text{target}}\).

\noindent\textbf{Search procedure:}
The resulting multidimensional, non-uniform interpolation is cast as a
search problem over the discrete grid \(P\):
LongRoPE performs coordinate-wise optimisation (or Bayesian search)
until convergence, then caches the \(\sigma_{j,k}\) table for use at
inference time.  
In practice \(K\!\ll\!L_{\text{target}}\) (e.g.\ anchors every 512
tokens), so memory overhead is negligible.

LongRoPE is more capable since it learns the parameter \(\sigma_{j,k}>0\) from the data and preserves original short-length quality, but it requires high-quality data for the adjustment process, and has a complex pipeline. 

\begin{table*}[t]
\centering
\small
\caption{Context extension at a glance. Tags: \textit{Context Extension} = frequency remapping \(h(\cdot)\) plus optional length/temperature scaling (e.g., YaRN); Arch.\ = architectural assists.}
\label{tab:ctx-ext-at-a-glance}
\resizebox{\textwidth}{!}{
\begin{tabular}{p{3.5cm}lllp{4.2cm}}
\toprule
Model & Max Len & Context Extension & Arch. & Datasets for long-context \\
\midrule
Nemotron-4 15B~\cite{parmar2024nemotron} & 4k & RoPE (native) & — & N/A \\
Gemma 1 (7B) & 8k & RoPE (native) & — & N/A \\
InternLM2~\cite{cai2024internlm2} & 32k & ABF & — & Book, patent, paper, CC. \\
MiniCPM-128K~\cite{hu2024minicpm} & 128k & ABF + NTK-aware & — & — \\
Phi-3-Mini~\cite{abdin2024phi3} & 128k & LongRoPE~\cite{ding2024longrope} & BlocksparseAttn & — \\
DeepSeek-V2~\cite{liu2024deepseek} & 128k & YaRN & MLA, MoE & — \\
Zamba-v1~\cite{glorioso2024zamba} & 4k & RoPE (native) & Hybrid ssm/global 6:1 & — \\
MAP-Neo~\cite{Zhang2024MAPNeoHC} & 8k & RoPE (native) & — & — \\
Nemotron-4 340B~\cite{adler2024nemotron} & 4k & RoPE (native) & — & N/A \\
AFM~\cite{gunter2024apple} & 32k & ABF & — & — \\
Llama-3.1~\cite{Dubey2024TheL3} & 128k & ABF + YaRN~\cite{peng2023yarn} & — & — \\
Gemma 2~\cite{team2024gemma} & 8k & RoPE (native) & Interleave local/global 1:1 & N/A \\
Granite 3.0~\cite{granite2024granite} & 4k & RoPE (native) & — & N/A \\
Hunyuan-Large~\cite{sun2024hunyuan} & 256k & ABF & MoE & 25\% Book, Code \\
Yi-Lightning~\cite{wake2024yi} & 64k & ABF & Interleave local/global 3:1 & — \\
Qwen-2.5/2.5-Turbo~\cite{Yang2024Qwen25TR} & 128k/1M & ABF +YaRN +DCA  & SparseAttn (Minference) & — \\
DeepSeek-V3~\cite{DeepSeekAI2024DeepSeekV3TR} & 128k & YaRN & MLA, MoE & — \\
OLMo-2~\cite{olmo20242} & 4k & RoPE (native) & — & N/A \\
MiniMax-01~\cite{li2025minimax} & 4M & ABF & Interleave LightningAttn/global 7:1 & +10\% long QA late-stage \\
Qwen-2.5-1M~\cite{yang2025qwen2} & 1M & ABF +YaRN +DCA & SparseAttn (Minference) & 40\% long +synthetic tasks \\
SmolLM2~\cite{allal2025smollm2} & 8k & ABF & — & 40\% long data mix \\
Gemma 3~\cite{team2025gemma} & 128k & ABF + PI & Interleave local/global 5:1 & — \\
Llama-4 Scout~\cite{meta2025llama4} & 10M & NoPE/iRoPE + temp scale & Interleave local/global & — \\
Phi-4~\cite{abdin2024phi} & 16k & ABF & — & +30\% curated long data \\
MiMo-7B~\cite{xiaomi2025mimo} & 32k & ABF & — & — \\
Qwen-3~\cite{yang2025qwen3} & 128k & ABF +YaRN +DCA & — & 75\% long data \\
PANGU PRO MoE~\cite{tang2025pangu} & 32k & — & MoE & STEM/code-heavy later stages \\
MiniCPM-4 8B~\cite{team2025minicpm4} & 128k & LongRoPE +YaRN & SparseAttn (InfLLM v2) & — \\
Hunyuan-A13B & 256k & NTK-aware & MoE & — \\
SmolLM3~\cite{bakouch2025smollm3} & 128k & ABF +YaRN & Interleave NoPE & math, code, reasoning-heavy \\
\bottomrule
\end{tabular}
}
\end{table*}

\subsubsection{A Unified View of Context Extension}
\label{sec:unified-context-ext}

For each rotary sub-plane \(j=1,\dots,\tfrac{d}{2}\) with base
frequency \(\theta_j\) (wavelength \(\lambda_j=2\pi/\theta_j\)),
and scale factor \(s=L_{\text{target}}/L_{\text{train}}>1\),
the following frequency maps \(h(\theta_j)\) instantiate the generic
template in Eq.~\eqref{eq:rope-extrapol-h}, as show in Table~\ref{tab:unified-h-theta}.

\medskip
\noindent\textbf{(i) Position Interpolation (PI).}
\begin{equation}
  h_{\text{PI}}(\theta_j)=\frac{\theta_j}{s}.
  \label{eq:h-pi}
\end{equation}

\medskip
\noindent\textbf{(ii) NTK-aware Scaling.}
\begin{equation}
  h_{\text{NTK}}(\theta_j)
  = \theta_j\,s^{-\tfrac{2(j-1)}{\,d-2\,}}.
  \label{eq:h-ntk}
\end{equation}

\medskip
\noindent\textbf{(iii) NTK-by-parts.}
Define the ratio
\begin{equation}
  r(j)=\frac{L_{\text{train}}}{\lambda_j}
      =\frac{L_{\text{train}}\theta_j}{2\pi},
  \label{eq:rj}
\end{equation}
and a ramp function with hyper-parameters \(\alpha<\beta\):
\begin{equation}
  \gamma(r)=
  \begin{cases}
    0, & r<\alpha,\\[4pt]
    1, & r>\beta,\\[4pt]
    \dfrac{r-\alpha}{\beta-\alpha}, & \text{otherwise}.
  \end{cases}
  \label{eq:gamma}
\end{equation}
Then
\begin{equation}
  h_{\text{NTP}}(\theta_j)
    = \bigl(1-\gamma(r(j))\bigr)\,\frac{\theta_j}{s}
      + \gamma(r(j))\,\theta_j.
  \label{eq:h-ntp-unified}
\end{equation}

\medskip
\noindent\textbf{(iv) YaRN.}
YaRN uses the NTK-by-parts frequency map unchanged:
\begin{equation}
  h_{\text{YaRN}}(\theta_j)=h_{\text{NTP}}(\theta_j).
  \label{eq:h-yarn}
\end{equation}
In addition, YaRN applies a length-dependent scaling to the resulting
queries/keys (or equivalently a softmax temperature):
\begin{equation}
  \alpha(s)=0.1\ln s + 1,
  \qquad
  t=\alpha(s)^{-2}.
  \label{eq:yarn-alpha-unified}
\end{equation}
(See Sec.~\ref{sec:yarn} for details.)

\medskip
\noindent\textbf{(v) LongRoPE.}
LongRoPE introduces non-uniform rescaling across RoPE dimensions and (optionally) across token positions. 
Let \(s_j \in [1,s]\) be a per-dimension rescale factor and let \(\tau\in\mathbb{N}\) denote the number of initial tokens that keep the original RoPE (no interpolation). 
Then the frequency map is
\begin{equation}
  h_{\text{LongRoPE}}(\theta_j;m)
  \;=\;
  \begin{cases}
    \theta_j, & m<\tau,\\[4pt]
    \dfrac{\theta_j}{\,s_j\,}, & m\ge \tau,
  \end{cases}
  \label{eq:h-longrope-unified}
\end{equation}
which recovers PI when \(s_j\equiv s\) and \(\tau=0\), and contains NTK/YaRN-like non-uniformity as special cases when \(s_j\) varies with \(j\). 
In our generic template \eqref{eq:rope-extrapol-h}, this corresponds to applying the rotation with an \(m\)- and \(j\)-dependent frequency,
\(
\mathbf R_{h_{\text{LongRoPE}}(\theta_j;m),\,m}.
\)
(See \cite{ding2024longrope} for the search-based procedure that selects \(\{s_j\}\) and \(\tau\).)

\begin{table*}[h]
  \centering
  \caption{Pre-training benchmarks by category.}
  \label{tab:std-benchmarks}
  \begin{tabular}{p{2.2cm}p{11cm}}
    \toprule
    \textbf{Task Category} & \textbf{Benchmarks} \\
    \midrule
    General & MMLU~\cite{hendrycks2021measuringmassivemultitasklanguage} (5-shot),
        MMLU-Pro~\cite{wang2024mmluprorobustchallengingmultitask} (5-shot, CoT),
        ARC ~\cite{clark2018thinksolvedquestionanswering} (0-shot), 
        HellaSwag~\cite{zellers2019hellaswagmachinereallyfinish} (0-shot),
        BBH~\cite{suzgun2022challengingbigbenchtaskschainofthought} (3-shot, CoT), 
        WinoGrande~\cite{sakaguchi2019winograndeadversarialwinogradschema} (0-shot),
        PIQA~\cite{bisk2019piqareasoningphysicalcommonsense} (0-shot) \\[2pt]
    \midrule
    Math \& STEM & GPQA~\cite{rein2023gpqagraduatelevelgoogleproofqa} (5-shot, CoT),
        MATH~\cite{hendrycks2021measuringmathematicalproblemsolving} (4-shot, CoT),
        GSM8K~\cite{cobbe2021gsm8k} (4-shot, CoT) \\[2pt]
    \midrule
    Coding & HumanEval~\cite{chen2021evaluating} (0-shot), 
        MBPP~\cite{austin2021programsynthesislargelanguage} (3-shot),  
        MultiPL-E~\cite{cassano2022multiplescalableextensibleapproach} (0-shot; Python, C++, Java, PHP, TypeScript, C\#, Bash, JavaScript) \\[2pt]
    \midrule
    Multilingual & MGSM~\cite{shi2022languagemodelsmultilingualchainofthought} (8-shot, CoT),
        Flores-101~\cite{goyal2021flores101evaluationbenchmarklowresource} (5-shot) \\[2pt]
    \midrule
    Long context & LongBench~\cite{bai2024longbenchbilingualmultitaskbenchmark}, 
        Ruler~\cite{hsieh2024rulerwhatsrealcontext} \\[2pt]
    \bottomrule
  \end{tabular}
\end{table*}

\subsection{Context Extension in Popular LLMs}
\label{sec:ctx-ext-popular}

Most systems scale context using three levers. Firstly, \emph{frequency remapping} of RoPE via \(h(\theta_j;s)\), such as ABF, PI, NTK, NTP and LongRoPE mentioned in Sec.~\ref{sec:unified-context-ext}. Secondly, optional \emph{temperature scaling} \(\alpha(s)\) at the attention logits, e.g., YaRN~\cite{peng2023yarn}. Thirdly, \emph{architecture assists} that reduce KV cost, e.g., the interleaved global and local attention, GQA, MLA, and MoE. 
In practice, windows are grown progressively \(4\text{k}\!\to\!32\text{k}\!\to\!128\text{k}\) with a mix of long and short tokens and occasional synthetic long-dependency tasks. We summarize the techniques used in many LLMs in Table~\ref{tab:ctx-ext-at-a-glance}.

\medskip

\noindent\textbf{No context extension (native RoPE).} 
The Nemotron-4 15B~\cite{parmar2024nemotron} uses native RoPE at 4K, and subsequent tuning reweights data but keeps the window unchanged. 
The Gemma 1 (7B) employs native RoPE at 8K. 
The Nemotron-4 340B is pretrained at 4K with native RoPE and does not include a long-context continuation phase. 
The Gemma 2~\cite{team2024gemma} fixed 8K with native RoPE, and it interleaves global and local layers roughly 1{:}1 to reduce KV cost.
The Granite 3.0~\cite{granite2024granite} family are trained at 4K with native RoPE. 
The OLMo 2~\cite{olmo20242} is a 4K model using native RoPE. 

\medskip

\noindent\textbf{Adaptive Base Frequency (ABF).}
The InternLM2~\cite{cai2024internlm2} extends to 32K via ABF~\cite{xiong2023effective} (RoPE base $50\mathrm{K}\!\to\!1\mathrm{M}$), GQA, and $\sim9\%$ mixed long sequences. Its training data includes book, patent, paper, and CC long sequences.
The Hunyuan-Large~\cite{sun2024hunyuan} uses staged 4K$\to$32K$\to$256K with ABF and a MoE backbone. About 25\% long natural data is included, and evaluations cover RULER/LV-Eval.
The Yi-Lightning~\cite{wake2024yi} uses progressive ABF from 8K to 64K. It combines sliding-window and full attention with an interleave of around 3{:}1.
The SmolLM2~\cite{allal2025smollm2} reaches 8K by increasing the RoPE base via ABF. It is trained with roughly 40\% long data.
The MiMo-7B~\cite{xiaomi2025mimo} follows three ABF stages to $32\mathrm{K}$ (RoPE $\theta:10\mathrm{K}\!\to\!64\mathrm{K}$). The data mix includes about 10\% synthetic math, code, and creative tasks to elicit longer-range behaviors.
The Gemma 3~\cite{team2025gemma} reaches $128\mathrm{K}$ with ABF on global layers plus position interpolation (PI). Global and local layers are interleaved at approximately $1{:}5$ for memory balance.
The Phi-4~\cite{abdin2024phi} reaches $16\mathrm{K}$ via ABF with a high RoPE base ($\sim\!250\mathrm{K}$). Later training up-weights natural and synthetic $>\!4\mathrm{K}$ data.
The SmolLM3~\cite{bakouch2025smollm3} scales $4\mathrm{K}\!\to\!64\mathrm{K}$ via ABF (with $\theta$ up to $1.5\mathrm{M}\!-\!5\mathrm{M}$) and selective NoPE for every 4th layer. Inference-time YaRN then lifts to $128\mathrm{K}$, focusing on math, code, and reasoning workloads. 

\medskip

\noindent\textbf{YaRN}.
DeepSeek-V2~\cite{liu2024deepseek} uses YaRN to scale $4\mathrm{K}\!\to\!128\mathrm{K}$. And the novel MLA and MoE design provide KV and compute efficiency.
DeepSeek-V3~\cite{DeepSeekAI2024DeepSeekV3TR} applies two YaRN phases with 1000 steps to extend the context length from ($4\mathrm{K}\!\to\!32\mathrm{K}\!\to\!128\mathrm{K}$). The MLA{+}MoE structure provides good efficiency, and it has strong long-context results.
Llama-3.1~\cite{Dubey2024TheL3} trains to $32\mathrm{K}$ then uses YaRN (with an adjusted RoPE base) to reach $128\mathrm{K}$. GQA reduces memory traffic for longer sequences.

\medskip

\noindent\textbf{DCA paired with ABF/YaRN at inference}.
The Qwen-2.5 / 2.5-Turbo~\cite{Yang2024Qwen25TR} uses ABF to extend their context to $128\mathrm{K}$, then uses inference-time DCA{+}YaRN to expand the window to $256\mathrm{K}+$. GQA and Minference-style sparse attention reduce KV cost at inference.
The Qwen-2.5-1M~\cite{yang2025qwen2} uses five ABF stages to extend their context to $256\mathrm{K}$ (with $\theta$ up to $10\mathrm{M}$), then uses DCA{+}YaRN push to $\sim\!1\mathrm{M}$ tokens. The data mix includes $\sim\!40\%$ long and synthetic tasks.
The Qwen-3~\cite{yang2025qwen3} uses ABF to support $32\mathrm{K}$ with a heavy long-data ratio ($\sim\!75\%$), then uses DCA{+}YaRN extend to $128\mathrm{K}$. The recipe retains ABF priors while scaling inference reach.

\medskip

\noindent\textbf{LongRoPE}.
The Phi-3 Mini (128K)~\cite{abdin2024phi3} achieves $128\mathrm{K}$ via LongRoPE~\cite{ding2024longrope} frequency remapping. The block-sparse attention also helps to manage memory and latency.
The MiniCPM-4 8B~\cite{team2025minicpm4} uses LongRoPE to support $32\mathrm{K}$ and uses inference-time YaRN to reach $128\mathrm{K}$. Sparse attention (InfLLM v2) further reduces KV pressure.

\medskip

\noindent\textbf{NTK-aware/PI/NoPE-family}.
The Hunyuan-A13B adopts NTK-aware scaling after a fast $4\mathrm{K}\!\to\!8\mathrm{K}$ stage. The temperature scale is $\alpha\!\approx\!50$ at $32\mathrm{K}$ and $\approx\!1000$ at $256\mathrm{K}$ with a MoE backbone.
The Llama-4 Scout~\cite{meta2025llama4} is pretrained at $256\mathrm{K}$ with iRoPE architecture, which has interleaved attention layers with no position embedding (NoPE). It uses inference-time temperature scaling. The reported inference scaling reaches $\sim\!10\mathrm{M}$ tokens.

\medskip

\noindent\textbf{Architecture-dominant}.
The MiniMax-01~\cite{li2025minimax} targets a $4\mathrm{M}$ context via a $7{:}1$ LightningAttn:full attention interleave. Its training upsamples long data and adds a late-stage QA mix to consolidate extremely long-span utility.
The PANGU PRO MoE~\cite{tang2025pangu} follows a three-phase training under $32\mathrm{K}$ on a MoE backbone. Later stages emphasize STEM and code to strengthen long-range reasoning.

\medskip

\noindent\textit{Takeaway.} Across families, a robust recipe pairs progressive ABF with a non-uniform \(h(\cdot)\) and a mild \(\alpha(s)\) schedule (e.g., YaRN), while architecture assists primarily unlock the memory budget required for long windows rather than replacing frequency remapping.


\begin{table*}[h]
  \centering
  \caption{Mid-Training uplift summary.}
  \label{tab:annealing uplift summary}
  \begin{tabular}{p{1.5cm}p{7.6cm}p{7.6cm}}
    \toprule
    \textbf{Model} & \textbf{Quantified Metric Uplift} & \textbf{Qualitative Impact / Purpose of Strategy} \\
    \midrule
    InternLM2     & Achieves substantial performance improvements in coding, reasoning, question answering, examinations, and long-context modeling. & \textbf{[Advanced Reasoning \& Adaptability]} Leveraging enriched datasets for coding, reasoning, QA, and exams, the model outperforms predecessors across 6 dimensions and 30 benchmarks, notably in long-context modeling and subjective evaluation. \\
    \midrule
    MiniCPM4      & Achieves comparable performance to Qwen3 using 22\% of tokens (8T vs 36T). MiniCPM4-0.5B surpasses Llama3.2-1B, Gemma3-1B. MiniCPM4-8B outperforms Gemma3-12B, Phi4-14B. & \textbf{[Scalable Efficiency]} Reduces training costs, improves performance \& efficiency, enables predictable scaling, critical for low-bit LLMs. \\
    \midrule
    DeepSeek-v2   & Achieves 88.9\% pass ratio on MiniF2F-test. Solves 47/658 PutnamBench problems. Pass rate can be improved significantly with CoT reasoning. & \textbf{[Advanced Reasoning \& Adaptability]} Overall training significantly enhances formal theorem proving capabilities. \\
    \midrule
    Zamba         & Base model (Phase 1) achieves Llama2 performance with 1T tokens (Llama2 uses 2T). Overall performance approaches leading ~7B transformers. & \textbf{[Advanced Reasoning \& Adaptability]} Improves adaptation, mitigates forgetting, enables fast adaptation from high-quality data, contributes to data efficiency. \\
    \midrule
    MAP-Neo       & Achieves rapid improvements and stabilization of code metrics during the decay phase, and enhances professional response generation. & \textbf{[Reliable Foundations]} Rectifies tokenizer flaws, enhances code generation, targeted remediation of specific deficiencies. \\
    \midrule
    Nemotron4     & Achieves a better learning from the data and a significant improvement in overall model quality. & \textbf{[Reliable Foundations]} Refines model, effectively integrates new data, contributes to overall polish and robustness. \\
    \midrule
    Llama3        & Annealing improved Llama 3 8B on GSM8k by 24.0\% and MATH by 6.4\%, but a negligible uplift for 405B model. & \textbf{[Scalable Efficiency]} Boosts performance for smaller models, ensures training stability, efficient method to judge value of domain-specific datasets. \\
    \midrule
    Gemma2        & Gemma2-9B shows up to 10\% improvement in some benchmarks compared to previous versions. & \textbf{[Scalable Efficiency]} Significantly improves model quality for smaller models, simulates training beyond available tokens, provides richer gradients. \\
    \midrule
    Hunyuan-Large & Achieves substantial improvements in downstream task performance and generalization ability. & \textbf{[Reliable Foundations]} Optimizes alignment pipeline, front-loads instruction-following, shapes model behavior, improves robustness. \\
    \midrule
    Yi-Lightning  & Achieves model capabilities enhancing and context length extending. & \textbf{[Scalable Efficiency]} Enables efficient training, deployment, inference, and high practical efficacy. Builds capabilities progressively from broad to specialized. \\
    \midrule
    DeepSeek-v3   & Achieves a remarkably stable training without irrecoverable loss spikes or rollbacks. & \textbf{[Scalable Efficiency]} Ensures robust and cost-effective training process, prevents costly failures, critical for large-scale training. \\
    \midrule
    OLMo2         & Mid-training improves 7B by 10.6 pts, 13B by 10.3 pts (avg performance). Specific uplifts on Arc Challenge (+7.2/3.3), MMLU (+3.9/4.1), GSM8K (+43.4/37.8), DROP (+20.1/21.1) for 7B and 13B respectively. & \textbf{[Reliable Foundations]} Dramatically improves performance across diverse benchmarks, enhances robustness, reduces experimentation cost. \\
    \midrule
    Phi-4         & Substantially surpasses teacher model on STEM-focused QA. Meets or exceeds Llama-3.1-405B on reasoning. Scores well above its weight-class on competition math. & \textbf{[Advanced Reasoning \& Adaptability]} Enables emergent reasoning capabilities, improves on teacher models, stabilizes learning during context extension. \\
    \midrule
    PanguProMoE   & Annealing phase can still bring about a large improvement in model performance, stabilize convergence, and reduce catastrophic forgetting. & \textbf{[Reliable Foundations]} Ensures robust generalization and effective specialization, systematically acquires diverse skills, prevents knowledge degradation. \\
    \midrule
    SmolLM3       & Achieves SOTA at 3B scale, and competitive performance with 4B models. Linear Blend retains robust CoT and instruction alignment while recovering top-tier performance on 128 K-token tasks. & \textbf{[Scalable Efficiency]} Maximizes performance from limited parameters, ensures robustness across capabilities, enables long-context robustness. \\
    \bottomrule
  \end{tabular}
\end{table*}

\subsection{Key paradigms and insights at a glance}
\label{sec:longcontext-takeaways}
\noindent{Frequency remapping and length/temperature scaling unify most methods.} Long-context schemes are well modeled by two independent controls:
(i) a \emph{frequency remapping} \(h(\theta_j;s)\) that modifies RoPE’s base angles in each rotary sub-plane (cf.\ Eq.~\eqref{eq:rope-extrapol-h}); and
(ii) an optional \emph{length/temperature scaling} \(\alpha(s)\) applied to queries/keys or softmax (e.g., YaRN in Eq.~\eqref{eq:yarn-alpha}).
PI, NTK-aware, NTK-by-parts, YaRN, and LongRoPE are all specific choices of \(h(\cdot)\) (and sometimes \(\alpha(\cdot)\)).

\noindent\textbf{Trade off between global reach and local fidelity.} Pure interpolation (e.g., PI, Eq.~\eqref{eq:h-pi}) stretches all frequencies equally but can blur local cues at long ranges; NTK-aware~(Eq.~\eqref{eq:h-ntk}) and NTK-by-parts~(Eq.~\eqref{eq:h-ntp}) spare high-frequency bands to preserve short-range structure; YaRN adds a gentle temperature schedule to stabilize optimization at large \(s\).


\noindent\textbf{Architecture assists matter but are orthogonal.} Interleaved global/local attention, GQA/MLA, and sparse attention primarily reduce KV-cache and memory cost; they \emph{enable} long sequences and pair well with frequency remapping, but do not replace it.

\noindent\textbf{Stage the data and schedule.} Effective recipes expand the window progressively (e.g., \(4\text{k}\!\rightarrow\!32\text{k}\!\rightarrow\!128\text{k}\)), anneal the learning rate, and mix long/short tokens (often 30–60\% long) with synthetic tasks (FIM, retrieval, reordering) to teach cross-document dependencies.


\noindent\textbf{Practical default.} For \(L_{\text{target}}\!\le\!128\text{k}\): set ABF to \(\theta\!\approx\!10^6\), 
apply YaRN \(\alpha(s)=0.1\ln s+1\). For \(\ge\!1\text{M}\): consider staged ABF (\(\theta\) up to \(10^7\!-\!10^8\)) and dimension/position-aware scaling (e.g., LongRoPE), while protecting short-context performance via short-length recovery.

\section{Evaluation}
\label{sec:evaluation}

\subsection{Standard Benchmarks}
\label{subsec:standard-benchmarks}
At the mid-training stage, model evaluation continues to rely predominantly on widely adopted standard benchmarks rather than task-specific or newly curated datasets. These benchmarks cover a broad spectrum of capabilities, including general knowledge and comprehension, reasoning and problem solving, mathematics and scientific knowledge, coding and software engineering, multilingual understanding, and long-context processing. Such coverage ensures comparability with prior work while highlighting progress across diverse domains. In our study, we report results on \textbf{15 representative benchmarks}, summarized in Table~\ref{tab:std-benchmarks}, which collectively reflect the core skill areas emphasized in large-scale LLM evaluation.

\subsection{Mid-Training Uplift}
The mid-training strategy has consistently been shown to enhance LLM performance, making it a central component of modern training practice. In this section, we summarize reported gains from a range of prominent models (see Table~\ref{tab:annealing uplift summary}). Our review suggests that the improvements attributed to mid-training fall into several major dimensions: including \textbf{(A)} Advanced Reasoning \& Adaptability, \textbf{(B)} Scalable Efficiency, \textbf{(C)} Reliable Foundations. To facilitate understanding, the table also categorizes these benefits by model, providing a structured comparison of how mid-training contributes across different training regimes.

\section{Future Work}
\label{sec:future}
Despite its growing adoption, \emph{mid-training} for LLMs remains more of an engineering art than a principled science. Below we outline key open challenges and sketch promising directions for future research.

\textbf{Dynamic Curriculum Design that adaptively adjust data composition.} 
A promising direction for mid-training is the development of dynamic curriculum strategies that adaptively adjust data composition based on the model’s evolving capabilities. Rather than relying on fixed sampling ratios, future approaches could monitor intermediate performance signals (e.g., loss sharpness, gradient variance, or benchmark outcomes) to guide the gradual transition from broad natural corpora toward specialized reasoning, coding, or multilingual tasks. Such adaptive curriculum would allow mid-training to better align data exposure with the model’s learning trajectory, improving efficiency while mitigating overfitting.

{\textbf{Theoretically grounded schedulers with adaptive mechanisms that co-evolve with scale and data.}}
{Learning rate schedulers remain a critical yet underexplored component in LLM training. Their interaction with other experimental parameters, such as batch size and model size, continues to be an active area of empirical and theoretical research. Future work should aim to unify these disparate empirical findings under a theoretical framework, and develop adaptive or scale-aware schedulers that generalize across model sizes and training regimes.}

\textbf{Long-context progress hinges on two fronts: fixing position-embedding OOD and cutting compute/latency.} Beyond RoPE tweaks, push toward general, learnable position functions and index warps, with robustness curricula and on-the-fly calibration. Beyond sparse attention, attack efficiency via KV-cache compression and quantization, token pruning/merging, hybrid attention-SSM stacks, retrieval-summarized contexts, and length-aware scheduling with compute controllers and MoE specialized by sequence length. We should also come up with better long-range datasets and metrics, and theory that ties spectral generalization and compute–length scaling laws—validated by concrete experiments on learned PEs, hybrid stacks, KV compression, retrieval summaries, and OOD calibration.

\section{Conclusion}
\label{sec:conclusion}




This paper presents a first-of-its-kind survey of LLM mid-training approaches. We introduce a taxonomy that categorizes existing methods into three key domains: data distribution, learning rate scheduling, and long-context extension. We summarize the main insights in each domain, providing a structured reference for researchers and practitioners. We also compile common evaluation benchmarks and reported gains, enabling a comparative view of how mid-training improves model performance. Finally, we identify open challenges and propose future research avenues, positioning mid-training as a central stage for shaping the next generation of large language models.

\begin{thebibliography}{100}
\providecommand{\url}[1]{#1}
\csname url@samestyle\endcsname
\providecommand{\newblock}{\relax}
\providecommand{\bibinfo}[2]{#2}
\providecommand{\BIBentrySTDinterwordspacing}{\spaceskip=0pt\relax}
\providecommand{\BIBentryALTinterwordstretchfactor}{4}
\providecommand{\BIBentryALTinterwordspacing}{\spaceskip=\fontdimen2\font plus
\BIBentryALTinterwordstretchfactor\fontdimen3\font minus \fontdimen4\font\relax}
\providecommand{\BIBforeignlanguage}[2]{{%
\expandafter\ifx\csname l@#1\endcsname\relax
\typeout{** WARNING: IEEEtran.bst: No hyphenation pattern has been}%
\typeout{** loaded for the language `#1'. Using the pattern for}%
\typeout{** the default language instead.}%
\else
\language=\csname l@#1\endcsname
\fi
#2}}
\providecommand{\BIBdecl}{\relax}
\BIBdecl

\bibitem{dubey2024llama}
A.~Dubey, A.~Jauhri, A.~Pandey \emph{et~al.}, ``The llama 3 herd of models,'' \emph{arXiv e-prints}, pp. arXiv--2407, 2024.

\bibitem{hu2024minicpm}
S.~Hu, Y.~Tu, X.~Han, C.~He \emph{et~al.}, ``Minicpm: Unveiling the potential of small language models with scalable training strategies,'' \emph{arXiv preprint arXiv:2404.06395}, 2024.

\bibitem{glorioso2024zamba}
P.~Glorioso, Q.~Anthony, Y.~Tokpanov \emph{et~al.}, ``Zamba: A compact 7b ssm hybrid model,'' \emph{arXiv preprint arXiv:2405.16712}, 2024.

\bibitem{abdin2024phi3}
M.~Abdin, J.~Aneja, H.~Awadalla, A.~Awadallah \emph{et~al.}, ``Phi-3 technical report: A highly capable language model locally on your phone,'' 2024.

\bibitem{bi2024deepseek}
X.~Bi, D.~Chen, G.~Chen \emph{et~al.}, ``Deepseek llm: Scaling open-source language models with longtermism,'' \emph{arXiv preprint arXiv:2401.02954}, 2024.

\bibitem{allal2025smollm2}
L.~B. Allal, A.~Lozhkov, E.~Bakouch, G.~M. Bl{\'a}zquez \emph{et~al.}, ``Smollm2: When smol goes big--data-centric training of a small language model,'' \emph{arXiv preprint arXiv:2502.02737}, 2025.

\bibitem{mccandlish2018empirical}
S.~McCandlish, J.~Kaplan, D.~Amodei, and O.~D. Team, ``An empirical model of large-batch training,'' \emph{arXiv preprint arXiv:1812.06162}, 2018.

\bibitem{wang2021eliminating}
X.~Wang, S.~Oh, and C.-H. Rhee, ``Eliminating sharp minima from sgd with truncated heavy-tailed noise,'' \emph{arXiv preprint arXiv:2102.04297}, 2021.

\bibitem{meng2023per}
X.~Meng, Y.~Cao, and D.~Zou, ``Per-example gradient regularization improves learning signals from noisy data,'' \emph{arXiv preprint arXiv:2303.17940}, 2023.

\bibitem{izmailov2022feature}
P.~Izmailov, P.~Kirichenko, N.~Gruver, and A.~G. Wilson, ``On feature learning in the presence of spurious correlations,'' \emph{Adv. Neural Inf. Process. Syst.}, vol.~35, pp. 38\,516--38\,532, 2022.

\bibitem{wang2025dump}
Z.~Wang, G.~Cui, Y.-J. Li, K.~Wan, and W.~Zhao, ``Dump: Automated distribution-level curriculum learning for rl-based llm post-training,'' \emph{arXiv preprint arXiv:2504.09710}, 2025.

\bibitem{jia2022sample}
Q.~Jia, Y.~Liu, H.~Tang, and K.~Q. Zhu, ``In-sample curriculum learning by sequence completion for natural language generation,'' \emph{arXiv preprint arXiv:2211.11297}, 2022.

\bibitem{du2024survey}
F.~Du, X.-J. Ma, J.-R. Yang \emph{et~al.}, ``A survey of llm datasets: From autoregressive model to ai chatbot,'' \emph{J. Comput. Sci. Technol.}, vol.~39, no.~3, pp. 542--566, 2024.

\bibitem{wu2025data}
O.~Wu, ``Data optimization for llms: A survey,'' \emph{Authorea Preprints}, 2025.

\bibitem{jin2023rethinking}
H.~Jin, W.~Wei, X.~Wang, W.~Zhang, and Y.~Wu, ``Rethinking learning rate tuning in the era of large language models,'' in \emph{2023 CogMI}.\hskip 1em plus 0.5em minus 0.4em\relax IEEE, 2023, pp. 112--121.

\bibitem{huang2023advancing}
Y.~Huang, J.~Xu, J.~Lai \emph{et~al.}, ``Advancing transformer architecture in long-context large language models: A comprehensive survey,'' \emph{arXiv preprint arXiv:2311.12351}, 2023.

\bibitem{liu2025comprehensive}
J.~Liu, D.~Zhu, Z.~Bai, Y.~He, H.~Liao, H.~Que \emph{et~al.}, ``A comprehensive survey on long context language modeling,'' \emph{arXiv preprint arXiv:2503.17407}, 2025.

\bibitem{pawar2024and}
S.~Pawar, S.~Tonmoy, S.~Zaman, V.~Jain, A.~Chadha, and A.~Das, ``The what, why, and how of context length extension techniques in large language models--a detailed survey,'' \emph{arXiv preprint arXiv:2401.07872}, 2024.

\bibitem{abdin2024phi}
M.~Abdin, J.~Aneja, H.~Behl \emph{et~al.}, ``Phi-4 technical report,'' \emph{arXiv preprint arXiv:2412.08905}, 2024.

\bibitem{NEURIPS2024_19e4ea30}
J.~Li, A.~Fang, G.~Smyrnis \emph{et~al.}, ``Datacomp-lm: In search of the next generation of training sets for language models,'' in \emph{Adv. Neural Inf. Process. Syst.}, vol.~37.\hskip 1em plus 0.5em minus 0.4em\relax Curran Associates, Inc., 2024, pp. 14\,200--14\,282.

\bibitem{penedo2024fineweb}
G.~Penedo, H.~Kydl{\'\i}{\v{c}}ek, A.~Lozhkov \emph{et~al.}, ``The fineweb datasets: Decanting the web for the finest text data at scale,'' \emph{Adv. Neural Inf. Process. Syst.}, vol.~37, pp. 30\,811--30\,849, 2024.

\bibitem{Su2024NemotronCCTC}
D.~Su, K.~Kong \emph{et~al.}, ``Nemotron-cc: Transforming common crawl into a refined long-horizon pretraining dataset,'' \emph{ArXiv}, vol. abs/2412.02595, 2024.

\bibitem{you1908does}
K.~You, M.~Long, J.~Wang, and M.~Jordan, ``How does learning rate decay help modern neural networks? arxiv 2019,'' \emph{arXiv preprint arXiv:1908.01878}, 1908.

\bibitem{gilmer2022loss}
J.~Gilmer, B.~Ghorbani \emph{et~al.}, ``A loss curvature perspective on training instabilities of deep learning models,'' in \emph{ICLR}, 2022.

\bibitem{kalra2024warmup}
D.~S. Kalra and M.~Barkeshli, ``Why warmup the learning rate? underlying mechanisms and improvements,'' \emph{Adv. Neural Inf. Process. Syst.}, vol.~37, pp. 111\,760--111\,801, 2024.

\bibitem{goyal2017accurate}
P.~Goyal, P.~Doll{\'a}r, R.~Girshick \emph{et~al.}, ``Accurate, large minibatch sgd: Training imagenet in 1 hour,'' \emph{arXiv preprint arXiv:1706.02677}, 2017.

\bibitem{loshchilov2016sgdr}
I.~Loshchilov and F.~Hutter, ``Sgdr: Stochastic gradient descent with warm restarts,'' \emph{arXiv preprint arXiv:1608.03983}, 2016.

\bibitem{shen2024power}
Y.~Shen, M.~Stallone, M.~Mishra, G.~Zhang \emph{et~al.}, ``Power scheduler: A batch size and token number agnostic learning rate scheduler,'' \emph{arXiv preprint arXiv:2408.13359}, 2024.

\bibitem{ibrahim2024simple}
A.~Ibrahim, B.~Th{\'e}rien, K.~Gupta \emph{et~al.}, ``Simple and scalable strategies to continually pre-train large language models,'' \emph{arXiv preprint arXiv:2403.08763}, 2024.

\bibitem{kaplan2020scaling}
J.~Kaplan, S.~McCandlish, T.~Henighan \emph{et~al.}, ``Scaling laws for neural language models,'' \emph{arXiv preprint arXiv:2001.08361}, 2020.

\bibitem{hoffmann2022training}
J.~Hoffmann, S.~Borgeaud, A.~Mensch \emph{et~al.}, ``Training compute-optimal large language models (2022),'' \emph{arXiv preprint arXiv:2203.15556}, 2022.

\bibitem{defazio2023optimal}
A.~Defazio, A.~Cutkosky, H.~Mehta, and K.~Mishchenko, ``Optimal linear decay learning rate schedules and further refinements,'' \emph{arXiv preprint arXiv:2310.07831}, 2023.

\bibitem{bergsma2025straight}
S.~Bergsma, N.~Dey, G.~Gosal, G.~Gray, D.~Soboleva, and J.~Hestness, ``Straight to zero: Why linearly decaying the learning rate to zero works best for llms,'' \emph{arXiv preprint arXiv:2502.15938}, 2025.

\bibitem{li2025predictable}
H.~Li, W.~Zheng, Q.~Wang, H.~Zhang, Z.~Wang \emph{et~al.}, ``Predictable scale: Part i--optimal hyperparameter scaling law in large language model pretraining,'' \emph{arXiv preprint arXiv:2503.04715}, 2025.

\bibitem{chen2023extending}
S.~Chen, S.~Wong, L.~Chen, and Y.~Tian, ``Extending context window of large language models via positional interpolation,'' \emph{arXiv preprint arXiv:2306.15595}, 2023.

\bibitem{bloc97_ntk_rope_2023}
bloc97, ``Ntk-aware scaled rope allows {LLaMA} models to have extended (8k+) context size without any fine-tuning and minimal perplexity degradation,'' Jun 2023, reddit post, r/LocalLLaMA.

\bibitem{peng2023yarn}
B.~Peng, J.~Quesnelle, H.~Fan, and E.~Shippole, ``Yarn: Efficient context window extension of large language models,'' \emph{arXiv preprint arXiv:2309.00071}, 2023.

\bibitem{xiong2023effective}
W.~Xiong, J.~Liu \emph{et~al.}, ``Effective long-context scaling of foundation models,'' \emph{arXiv preprint arXiv:2309.16039}, 2023.

\bibitem{soldaini2024dolma}
L.~Soldaini, R.~Kinney, A.~Bhagia \emph{et~al.}, ``Dolma: An open corpus of three trillion tokens for language model pretraining research,'' \emph{arXiv preprint arXiv:2402.00159}, 2024.

\bibitem{C4dataset}
allenai. (2021)

\bibitem{gao2020pile}
L.~Gao, S.~Biderman, S.~Black, L.~Golding \emph{et~al.}, ``The pile: An 800gb dataset of diverse text for language modeling,'' \emph{arXiv preprint arXiv:2101.00027}, 2020.

\bibitem{wikidump}
W.~Foundation. Wikimedia downloads.

\bibitem{clement2019usearxivdataset}
C.~B. Clement, M.~Bierbaum, K.~P. O'Keeffe, and A.~A. Alemi, ``On the use of arxiv as a dataset,'' 2019.

\bibitem{Kocetkov2022TheStack}
D.~Kocetkov, R.~Li, L.~Ben~Allal, J.~Li \emph{et~al.}, ``The stack: 3 tb of permissively licensed source code,'' \emph{Preprint}, 2022.

\bibitem{paster2023openwebmath}
K.~Paster, M.~D. Santos, Z.~Azerbayev, and J.~Ba, ``Openwebmath: An open dataset of high-quality mathematical web text,'' 2023.

\bibitem{SlimOrca}
W.~Lian, G.~Wang, B.~Goodson \emph{et~al.}, ``Slimorca: An open dataset of gpt-4 augmented flan reasoning traces, with verification,'' 2023.

\bibitem{wei2024magicoder}
Y.~Wei, Z.~Wang, J.~Liu, Y.~Ding, and L.~Zhang, ``Magicoder: Empowering code generation with {OSS}-instruct,'' in \emph{ICML}, 21--27 Jul 2024.

\bibitem{ding2023enhancing}
N.~Ding, Y.~Chen, B.~Xu, Y.~Qin, Z.~Zheng \emph{et~al.}, ``Enhancing chat language models by scaling high-quality instructional conversations,'' \emph{arXiv preprint arXiv:2305.14233}, 2023.

\bibitem{surge2024openbezoar}
C.~Dissanayake, L.~Lowe, S.~Gunasekara, and Y.~Ratnayake, ``Openbezoar: Small, cost-effective and open models trained on mixes of instruction data,'' 2024.

\bibitem{peS2o}
L.~Soldaini and K.~Lo, ``{peS2o (Pretraining Efficiently on S2ORC) Dataset},'' {Allen Institute for AI}, Tech. Rep., 2023, oDC-By, \url{https://github.com/allenai/pes2o}.

\bibitem{penedo2023refinedweb}
G.~Penedo, Q.~Malartic, D.~Hesslow, R.~Cojocaru \emph{et~al.}, ``The refinedweb dataset for falcon llm: outperforming curated corpora with web data, and web data only,'' \emph{arXiv preprint arXiv:2306.01116}, 2023.

\bibitem{luo2023wizardcoder}
Z.~Luo, C.~Xu, P.~Zhao \emph{et~al.}, ``Wizardcoder: Empowering code large language models with evol-instruct,'' 2023.

\bibitem{stackmathqa2024}
Y.~Zhang, ``Stackmathqa: A curated collection of 2 million mathematical questions and answers sourced from stack exchange,'' 2024.

\bibitem{OpenOrca}
W.~Lian, B.~Goodson, E.~Pentland, A.~Cook, C.~Vong, and "Teknium", ``Openorca: An open dataset of gpt augmented flan reasoning traces,'' \url{https://https://huggingface.co/datasets/Open-Orca/OpenOrca}, 2023.

\bibitem{benallal2024cosmopedia}
L.~Ben~Allal, A.~Lozhkov, G.~Penedo, T.~Wolf, and L.~von Werra, ``Cosmopedia,'' 2024.

\bibitem{Zhang2024MAPNeoHC}
G.~Zhang, S.~Qu \emph{et~al.}, ``Map-neo: Highly capable and transparent bilingual large language model series,'' \emph{ArXiv}, vol. abs/2405.19327, 2024.

\bibitem{US-PD-Books}
``Us-pd-books: Us public domain books (english),'' \url{https://huggingface.co/datasets/storytracer/US-PD-Books}, 2024.

\bibitem{lozhkov2024starcoder}
A.~Lozhkov, R.~Li, L.~B. Allal, F.~Cassano \emph{et~al.}, ``Starcoder 2 and the stack v2: The next generation,'' 2024.

\bibitem{zhang2025autonomous}
Y.~Zhang, Y.~Luo, Y.~Yuan, and A.~C.-C. Yao, ``Autonomous data selection with zero-shot generative classifiers for mathematical texts,'' \emph{ACL Findings}, 2025.

\bibitem{BioInstructQA}
BioMistral, ``Bioinstructqa,'' 2024.

\bibitem{yu2024llasmol}
B.~Yu, F.~N. Baker, Z.~Chen \emph{et~al.}, ``Lla{SM}ol: Advancing large language models for chemistry with a large-scale, comprehensive, high-quality instruction tuning dataset,'' in \emph{COLM}, 2024.

\bibitem{chen2024agent}
Z.~Chen, K.~Liu, Q.~Wang, J.~Liu \emph{et~al.}, ``Agent-flan: Designing data and methods of effective agent tuning for large language models,'' \emph{arXiv preprint arXiv:2403.12881}, 2024.

\bibitem{olmo20242}
T.~OLMo, P.~Walsh, L.~Soldaini, D.~Groeneveld \emph{et~al.}, ``2 olmo 2 furious,'' \emph{arXiv preprint arXiv:2501.00656}, 2024.

\bibitem{husain2019codesearchnet}
H.~Husain, H.-H. Wu \emph{et~al.}, ``Codesearchnet challenge: Evaluating the state of semantic code search,'' \emph{arXiv preprint arXiv:1909.09436}, 2019.

\bibitem{yu2023metamath}
L.~Yu, W.~Jiang, H.~Shi, J.~Yu \emph{et~al.}, ``Metamath: Bootstrap your own mathematical questions for large language models,'' \emph{arXiv preprint arXiv:2309.12284}, 2023.

\bibitem{cobbe2021gsm8k}
K.~Cobbe, V.~Kosaraju, M.~Bavarian \emph{et~al.}, ``Training verifiers to solve math word problems,'' \emph{arXiv preprint arXiv:2110.14168}, 2021.

\bibitem{together2023redpajama}
M.~Weber, D.~Y. Fu \emph{et~al.}, ``Redpajama: an open dataset for training large language models,'' \emph{NeurIPS Datasets and Benchmarks Track}, 2024.

\bibitem{han2024infimmwebmath40badvancingmultimodalpretraining}
X.~Han, Y.~Jian, X.~Hu, H.~Liu \emph{et~al.}, ``Infimm-webmath-40b: Advancing multimodal pre-training for enhanced mathematical reasoning,'' 2024.

\bibitem{li2023mugglemath}
C.~Li, Z.~Yuan, H.~Yuan, G.~Dong \emph{et~al.}, ``Mugglemath: Assessing the impact of query and response augmentation on math reasoning,'' \emph{arXiv preprint arXiv:2310.05506}, 2023.

\bibitem{benallal2024smollmcorpus}
L.~Ben~Allal, A.~Lozhkov, G.~Penedo, T.~Wolf, and L.~von Werra, ``Smollm-corpus,'' 2024.

\bibitem{neelakantan2015adding}
A.~Neelakantan, L.~Vilnis, Q.~V. Le, I.~Sutskever \emph{et~al.}, ``Adding gradient noise improves learning for very deep networks,'' \emph{arXiv preprint arXiv:1511.06807}, 2015.

\bibitem{tishby2015deep}
N.~Tishby and N.~Zaslavsky, ``Deep learning and the information bottleneck principle,'' in \emph{IEEE ITW}, 2015, pp. 1--5.

\bibitem{alemi2016deep}
A.~A. Alemi, I.~Fischer, J.~V. Dillon, and K.~Murphy, ``Deep variational information bottleneck,'' \emph{arXiv preprint arXiv:1612.00410}, 2016.

\bibitem{weingarten2025supervised}
N.~Z. Weingarten, Z.~Yakhini, M.~Butman, and R.~Bustin, ``The supervised information bottleneck,'' \emph{Entropy}, vol.~27, no.~5, p. 452, 2025.

\bibitem{nair2024curriculum}
M.~Na{\"\i}r, K.~Yamani, L.~S. Lhadj, and R.~Baghdadi, ``Curriculum learning for small code language models,'' \emph{arXiv preprint arXiv:2407.10194}, 2024.

\bibitem{chaudhry2024data}
S.~Chaudhry and A.~Sharma, ``Data distribution-based curriculum learning,'' \emph{IEEE Access}, 2024.

\bibitem{sun2024hunyuan}
X.~Sun, Y.~Chen \emph{et~al.}, ``Hunyuan-large: An open-source moe model with 52 billion activated parameters by tencent,'' \emph{arXiv preprint arXiv:2411.02265}, 2024.

\bibitem{Dubey2024TheL3}
A.~Dubey, A.~Jauhri \emph{et~al.}, ``The llama 3 herd of models,'' \emph{ArXiv}, vol. abs/2407.21783, 2024.

\bibitem{nguyen2023culturax}
T.~Nguyen, C.~Van~Nguyen, V.~D. Lai, H.~Man, N.~T. Ngo, F.~Dernoncourt, R.~A. Rossi, and T.~H. Nguyen, ``Culturax: A cleaned, enormous, and multilingual dataset for large language models in 167 languages,'' \emph{arXiv preprint arXiv:2309.09400}, 2023.

\bibitem{shen2023slimpajama}
Z.~Shen, T.~Tao \emph{et~al.}, ``Slimpajama-dc: Understanding data combinations for llm training,'' \emph{arXiv preprint arXiv:2309.10818}, 2023.

\bibitem{gunter2024apple}
T.~Gunter, Z.~Wang, C.~Wang \emph{et~al.}, ``Apple intelligence foundation language models,'' \emph{arXiv preprint arXiv:2407.21075}, 2024.

\bibitem{chen2025minimax}
A.~Chen, A.~Li, B.~Gong, B.~Jiang \emph{et~al.}, ``Minimax-m1: Scaling test-time compute efficiently with lightning attention,'' \emph{arXiv preprint arXiv:2506.13585}, 2025.

\bibitem{yang2025qwen3}
A.~Yang, A.~Li, B.~Yang, B.~Zhang \emph{et~al.}, ``Qwen3 technical report,'' \emph{arXiv preprint arXiv:2505.09388}, 2025.

\bibitem{han2024infimm}
X.~Han, Y.~Jian \emph{et~al.}, ``Infimm-webmath-40b: Advancing multimodal pre-training for enhanced mathematical reasoning,'' \emph{arXiv preprint arXiv:2409.12568}, 2024.

\bibitem{huang2024opencoder}
S.~Huang, T.~Cheng \emph{et~al.}, ``Opencoder: The open cookbook for top-tier code large language models,'' \emph{arXiv preprint arXiv:2411.04905}, 2024.

\bibitem{adler2024nemotron}
B.~Adler, N.~Agarwal, A.~Aithal, D.~H. Anh \emph{et~al.}, ``Nemotron-4 340b technical report,'' \emph{arXiv preprint arXiv:2406.11704}, 2024.

\bibitem{tang2025pangu}
Y.~Tang, X.~Li, F.~Liu, W.~Guo \emph{et~al.}, ``Pangu pro moe: Mixture of grouped experts for efficient sparsity,'' \emph{arXiv preprint arXiv:2505.21411}, 2025.

\bibitem{longpre2023flan}
S.~Longpre, L.~Hou, T.~Vu, A.~Webson \emph{et~al.}, ``The flan collection: Designing data and methods for effective instruction tuning,'' in \emph{ICML}.\hskip 1em plus 0.5em minus 0.4em\relax PMLR, 2023, pp. 22\,631--22\,648.

\bibitem{gao2024train}
T.~Gao, A.~Wettig, H.~Yen, and D.~Chen, ``How to train long-context language models (effectively),'' \emph{arXiv preprint arXiv:2410.02660}, 2024.

\bibitem{sprague2024cot}
Z.~Sprague, F.~Yin, J.~D. Rodriguez, D.~Jiang \emph{et~al.}, ``To cot or not to cot? chain-of-thought helps mainly on math and symbolic reasoning,'' \emph{arXiv preprint arXiv:2409.12183}, 2024.

\bibitem{DeepSeekAI2024DeepSeekV3TR}
DeepSeek-AI, A.~Liu \emph{et~al.}, ``Deepseek-v3 technical report,'' \emph{ArXiv}, vol. abs/2412.19437, 2024.

\bibitem{zhu2024deepseek}
Q.~Zhu, D.~Guo, Z.~Shao, D.~Yang, P.~Wang \emph{et~al.}, ``Deepseek-coder-v2: Breaking the barrier of closed-source models in code intelligence,'' \emph{arXiv preprint arXiv:2406.11931}, 2024.

\bibitem{li2023starcoder}
R.~Li, L.~B. Allal, Y.~Zi, N.~Muennighoff \emph{et~al.}, ``Starcoder: may the source be with you!'' \emph{arXiv preprint arXiv:2305.06161}, 2023.

\bibitem{wang2023codet5+}
Y.~Wang, H.~Le, A.~D. Gotmare \emph{et~al.}, ``Codet5+: Open code large language models for code understanding and generation,'' \emph{arXiv preprint arXiv:2305.07922}, 2023.

\bibitem{granite2024granite}
I.~Granite~Team, ``Granite 3.0 language models,'' \emph{URL: https://github. com/ibm-granite/granite-3.0-language-models}, 2024.

\bibitem{wake2024yi}
A.~Wake, B.~Chen, C.~Lv, C.~Li, C.~Huang \emph{et~al.}, ``Yi-lightning technical report,'' \emph{arXiv preprint arXiv:2412.01253}, 2024.

\bibitem{xiaomi2025mimo}
L.~Xiaomi, B.~Xia, B.~Shen, D.~Zhu \emph{et~al.}, ``Mimo: Unlocking the reasoning potential of language model--from pretraining to posttraining,'' \emph{arXiv preprint arXiv:2505.07608}, 2025.

\bibitem{parmar2024nemotron}
J.~Parmar, S.~Prabhumoye, J.~Jennings \emph{et~al.}, ``Nemotron-4 15b technical report,'' \emph{arXiv preprint arXiv:2402.16819}, 2024.

\bibitem{wang2025ultra}
Y.~Wang, Z.~Fu, J.~Cai, P.~Tang \emph{et~al.}, ``Ultra-fineweb: Efficient data filtering and verification for high-quality llm training data,'' \emph{arXiv preprint arXiv:2505.05427}, 2025.

\bibitem{springer2025overtrained}
J.~M. Springer, S.~Goyal \emph{et~al.}, ``Overtrained language models are harder to fine-tune,'' \emph{arXiv preprint arXiv:2503.19206}, 2025.

\bibitem{vaswani2017attention}
A.~Vaswani, N.~Shazeer, N.~Parmar, J.~Uszkoreit, L.~Jones, A.~N. Gomez, {\L}.~Kaiser, and I.~Polosukhin, ``Attention is all you need,'' \emph{Adv. Neural Inf. Process. Syst.}, vol.~30, 2017.

\bibitem{brown2020language}
T.~Brown, B.~Mann, N.~Ryder, M.~Subbiah \emph{et~al.}, ``Language models are few-shot learners,'' \emph{Adv. Neural Inf. Process. Syst.}, vol.~33, pp. 1877--1901, 2020.

\bibitem{rae2021scaling}
J.~W. Rae, S.~Borgeaud, T.~Cai \emph{et~al.}, ``Scaling language models: Methods, analysis \& insights from training gopher,'' \emph{arXiv preprint arXiv:2112.11446}, 2021.

\bibitem{li2019exponential}
Z.~Li and S.~Arora, ``An exponential learning rate schedule for deep learning,'' \emph{arXiv preprint arXiv:1910.07454}, 2019.

\bibitem{iyer2023wide}
N.~Iyer, V.~Thejas, N.~Kwatra, R.~Ramjee, and M.~Sivathanu, ``Wide-minima density hypothesis and the explore-exploit learning rate schedule,'' \emph{J. Mach. Learn. Res}, vol.~24, no.~65, pp. 1--37, 2023.

\bibitem{smith2017cyclical}
L.~N. Smith, ``Cyclical learning rates for training neural networks,'' in \emph{IEEE WACV}.\hskip 1em plus 0.5em minus 0.4em\relax IEEE, 2017, pp. 464--472.

\bibitem{devlin2019bert}
J.~Devlin, M.-W. Chang, K.~Lee, and K.~Toutanova, ``Bert: Pre-training of deep bidirectional transformers for language understanding,'' in \emph{NAACL-HLT}, 2019, pp. 4171--4186.

\bibitem{chowdhery2023palm}
A.~Chowdhery, S.~Narang, J.~Devlin \emph{et~al.}, ``Palm: Scaling language modeling with pathways,'' \emph{J. Mach. Learn. Res}, vol.~24, no. 240, pp. 1--113, 2023.

\bibitem{zhang2022opt}
S.~Zhang, S.~Roller, N.~Goyal, M.~Artetxe \emph{et~al.}, ``Opt: Open pre-trained transformer language models,'' \emph{arXiv preprint arXiv:2205.01068}, 2022.

\bibitem{workshop2022bloom}
B.~Workshop, T.~L. Scao, A.~Fan \emph{et~al.}, ``Bloom: A 176b-parameter open-access multilingual language model,'' \emph{arXiv preprint arXiv:2211.05100}, 2022.

\bibitem{kreps2022all}
S.~Kreps, R.~M. McCain, and M.~Brundage, ``All the news that’s fit to fabricate: Ai-generated text as a tool of media misinformation,'' \emph{J EXP POLIT SCI.}, vol.~9, no.~1, pp. 104--117, 2022.

\bibitem{shazeer2018adafactor}
N.~Shazeer and M.~Stern, ``Adafactor: Adaptive learning rates with sublinear memory cost,'' in \emph{ICML}, 2018, pp. 4596--4604.

\bibitem{touvron2023llama}
H.~Touvron, T.~Lavril, G.~Izacard, X.~Martinet \emph{et~al.}, ``Llama: Open and efficient foundation language models,'' \emph{arXiv preprint arXiv:2302.13971}, 2023.

\bibitem{touvron2023llama2}
H.~Touvron, L.~Martin, K.~Stone \emph{et~al.}, ``Llama 2: Open foundation and fine-tuned chat models,'' \emph{arXiv preprint arXiv:2307.09288}, 2023.

\bibitem{bai2023qwen}
J.~Bai, S.~Bai, Y.~Chu, Z.~Cui, K.~Dang \emph{et~al.}, ``Qwen technical report,'' \emph{arXiv preprint arXiv:2309.16609}, 2023.

\bibitem{qwen2025qwen25technicalreport}
Qwen, :, A.~Yang, B.~Yang, B.~Zhang, B.~Hui \emph{et~al.}, ``Qwen2.5 technical report,'' 2025.

\bibitem{liu2024deepseek}
A.~Liu, B.~Feng \emph{et~al.}, ``Deepseek-v2: A strong, economical, and efficient mixture-of-experts language model,'' \emph{arXiv preprint arXiv:2405.04434}, 2024.

\bibitem{liu2024deepseekv3}
A.~Liu, B.~Feng, B.~Xue, B.~Wang \emph{et~al.}, ``Deepseek-v3 technical report,'' \emph{arXiv preprint arXiv:2412.19437}, 2024.

\bibitem{cai2024internlm2}
Z.~Cai, M.~Cao, H.~Chen, K.~Chen, K.~Chen, X.~Chen, X.~Chen, Z.~Chen, Z.~Chen, P.~Chu \emph{et~al.}, ``Internlm2 technical report,'' \emph{arXiv preprint arXiv:2403.17297}, 2024.

\bibitem{almazrouei2023falcon}
E.~Almazrouei, H.~Alobeidli \emph{et~al.}, ``The falcon series of open language models,'' \emph{arXiv preprint arXiv:2311.16867}, 2023.

\bibitem{raffel2020exploring}
C.~Raffel, N.~Shazeer, A.~Roberts, K.~Lee \emph{et~al.}, ``Exploring the limits of transfer learning with a unified text-to-text transformer,'' \emph{J. Mach. Learn. Res}, vol.~21, no. 140, pp. 1--67, 2020.

\bibitem{xue2024openmoe}
F.~Xue, Z.~Zheng, Y.~Fu, J.~Ni, Z.~Zheng, W.~Zhou, and Y.~You, ``Openmoe: An early effort on open mixture-of-experts language models,'' \emph{arXiv preprint arXiv:2402.01739}, 2024.

\bibitem{hu2024yulan}
Y.~Hu, H.~Song, J.~Deng \emph{et~al.}, ``Yulan-mini: An open data-efficient language model,'' \emph{arXiv preprint arXiv:2412.17743}, 2024.

\bibitem{li2025minimax}
A.~Li, B.~Gong, B.~Yang, B.~Shan \emph{et~al.}, ``Minimax-01: Scaling foundation models with lightning attention,'' \emph{arXiv preprint arXiv:2501.08313}, 2025.

\bibitem{nijkamp2025xgen}
E.~Nijkamp, B.~Pang, E.~Pakhomov, A.~Gokul, J.~Qu, S.~Savarese, Y.~Zhou, and C.~Xiong, ``xgen-small technical report,'' \emph{arXiv preprint arXiv:2505.06496}, 2025.

\bibitem{tancik2020fourier}
M.~Tancik, P.~Srinivasan \emph{et~al.}, ``Fourier features let networks learn high frequency functions in low dimensional domains,'' \emph{Adv. Neural Inf. Process. Syst.}, vol.~33, pp. 7537--7547, 2020.

\bibitem{ding2024longrope}
Y.~Ding, L.~L. Zhang, C.~Zhang, Y.~Xu, N.~Shang, J.~Xu, F.~Yang, and M.~Yang, ``Longrope: Extending llm context window beyond 2 million tokens,'' \emph{arXiv preprint arXiv:2402.13753}, 2024.

\bibitem{team2024gemma}
G.~Team, M.~Riviere, S.~Pathak, P.~G. Sessa \emph{et~al.}, ``Gemma 2: Improving open language models at a practical size,'' \emph{arXiv preprint arXiv:2408.00118}, 2024.

\bibitem{Yang2024Qwen25TR}
Q.~A. Yang, B.~Yang \emph{et~al.}, ``Qwen2.5 technical report,'' \emph{ArXiv}, vol. abs/2412.15115, 2024.

\bibitem{yang2025qwen2}
A.~Yang, B.~Yu \emph{et~al.}, ``Qwen2. 5-1m technical report,'' \emph{arXiv preprint arXiv:2501.15383}, 2025.

\bibitem{team2025gemma}
Gemma, A.~Kamath, J.~Ferret \emph{et~al.}, ``Gemma 3 technical report,'' \emph{arXiv preprint arXiv:2503.19786}, 2025.

\bibitem{meta2025llama4}
{Meta AI}, ``{Llama 4 Scout 17B-16E Instruct},'' \url{https://huggingface.co/meta-llama/Llama-4-Scout-17B-16E-Instruct}, 2025, model card. Version effective 5 Apr 2025. Accessed 3 Jul 2025.

\bibitem{team2025minicpm4}
M.~Team, C.~Xiao, Y.~Li, X.~Han \emph{et~al.}, ``Minicpm4: Ultra-efficient llms on end devices,'' \emph{arXiv preprint arXiv:2506.07900}, 2025.

\bibitem{bakouch2025smollm3}
E.~Bakouch, C.~M. Pati{\~n}o \emph{et~al.}, ``Smollm3: smol, multilingual, long-context reasoner,'' Hugging Face Blog, July 2025.

\bibitem{hendrycks2021measuringmassivemultitasklanguage}
D.~Hendrycks, C.~Burns, S.~Basart, A.~Zou, M.~Mazeika, D.~Song, and J.~Steinhardt, ``Measuring massive multitask language understanding,'' 2021.

\bibitem{wang2024mmluprorobustchallengingmultitask}
Y.~Wang, X.~Ma \emph{et~al.}, ``Mmlu-pro: A more robust and challenging multi-task language understanding benchmark,'' 2024.

\bibitem{clark2018thinksolvedquestionanswering}
P.~Clark, I.~Cowhey, O.~Etzioni, T.~Khot, A.~Sabharwal, C.~Schoenick, and O.~Tafjord, ``Think you have solved question answering? try arc, the ai2 reasoning challenge,'' 2018.

\bibitem{zellers2019hellaswagmachinereallyfinish}
R.~Zellers, A.~Holtzman, Y.~Bisk, A.~Farhadi, and Y.~Choi, ``Hellaswag: Can a machine really finish your sentence?'' 2019.

\bibitem{suzgun2022challengingbigbenchtaskschainofthought}
M.~Suzgun, N.~Scales, N.~Schärli \emph{et~al.}, ``Challenging big-bench tasks and whether chain-of-thought can solve them,'' 2022.

\bibitem{sakaguchi2019winograndeadversarialwinogradschema}
K.~Sakaguchi, R.~L. Bras \emph{et~al.}, ``Winogrande: An adversarial winograd schema challenge at scale,'' 2019.

\bibitem{bisk2019piqareasoningphysicalcommonsense}
Y.~Bisk, R.~Zellers \emph{et~al.}, ``Piqa: Reasoning about physical commonsense in natural language,'' 2019.

\bibitem{rein2023gpqagraduatelevelgoogleproofqa}
D.~Rein, B.~L. Hou, A.~C. Stickland \emph{et~al.}, ``Gpqa: A graduate-level google-proof q\&a benchmark,'' 2023.

\bibitem{hendrycks2021measuringmathematicalproblemsolving}
D.~Hendrycks, C.~Burns, S.~Kadavath, A.~Arora, S.~Basart, E.~Tang, D.~Song, and J.~Steinhardt, ``Measuring mathematical problem solving with the math dataset,'' 2021.

\bibitem{chen2021evaluating}
M.~Chen, J.~Tworek, H.~Jun, Q.~Yuan, H.~P. D.~O. Pinto \emph{et~al.}, ``Evaluating large language models trained on code,'' \emph{arXiv preprint arXiv:2107.03374}, 2021.

\bibitem{austin2021programsynthesislargelanguage}
J.~Austin, A.~Odena, M.~Nye \emph{et~al.}, ``Program synthesis with large language models,'' 2021.

\bibitem{cassano2022multiplescalableextensibleapproach}
F.~Cassano, J.~Gouwar \emph{et~al.}, ``Multipl-e: A scalable and extensible approach to benchmarking neural code generation,'' 2022.

\bibitem{shi2022languagemodelsmultilingualchainofthought}
F.~Shi, M.~Suzgun, M.~Freitag \emph{et~al.}, ``Language models are multilingual chain-of-thought reasoners,'' 2022.

\bibitem{goyal2021flores101evaluationbenchmarklowresource}
N.~Goyal, C.~Gao, V.~Chaudhary, P.-J. Chen \emph{et~al.}, ``The flores-101 evaluation benchmark for low-resource and multilingual machine translation,'' 2021.

\bibitem{bai2024longbenchbilingualmultitaskbenchmark}
Y.~Bai, X.~Lv, J.~Zhang \emph{et~al.}, ``Longbench: A bilingual, multitask benchmark for long context understanding,'' 2024.

\bibitem{hsieh2024rulerwhatsrealcontext}
C.-P. Hsieh, S.~Sun, S.~Kriman, S.~Acharya \emph{et~al.}, ``Ruler: What's the real context size of your long-context language models?'' 2024.

\end{thebibliography}

\end{document}